%% file: main.tex
\documentclass{article} % For LaTeX2e
\usepackage{iclr2025_conference,times}

\usepackage{wrapfig}
\usepackage{lipsum} % 用于生成示例文字
\usepackage{wrapfig}

\input{math_commands.tex}

\usepackage{enumerate}
\usepackage{hyperref}
\usepackage{url}
\usepackage{xcolor}
\usepackage{microtype}
\usepackage{graphicx}
\usepackage{marvosym}
\usepackage{tcolorbox}
\usepackage{subfigure}
\usepackage{geometry}
\usepackage{subfloat}
\usepackage{listings}
\usepackage{xcolor}
\usepackage{booktabs} % for professional tables
\usepackage{float}
\usepackage{amssymb}
\usepackage{color}
\usepackage{mathtools}
\usepackage{amsthm}
\usepackage{multirow}
\usepackage{cleveref}
\usepackage{caption}
\usepackage{mathtools}
\usepackage{amsthm}
\usepackage{amsmath}
\usepackage{algorithm} 
\usepackage{algorithmicx}
\usepackage{algpseudocode}
\usepackage{setspace} 
\usepackage{wrapfig}
\usepackage{cleveref}
\usepackage{diagbox}
\usepackage{xcolor}
\usepackage{svg}
\usepackage{adjustbox} % 用于调整代码框宽度
\usepackage{minted}
\usepackage{fancyvrb}

\definecolor{uclablue}{rgb}{0.15, 0.45, 0.68}
\hypersetup{
    breaklinks,
    citecolor=uclablue,
    colorlinks=true,
}

\NewDocumentCommand{\xx}
{ mO{} }{\textcolor{blue}{\textsuperscript{\textit{todo}}\textsf{\textbf{\small[#1]}}}}
\usepackage{xspace}
\newcommand{\github}{\raisebox{-1.5pt}{\includegraphics[height=1.05em]{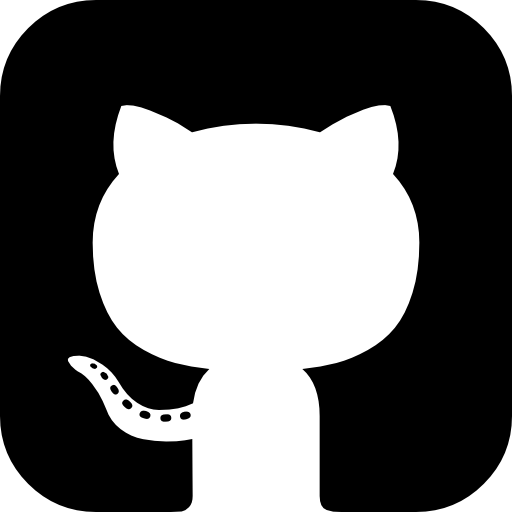}}\xspace}
\newcommand{\dataset}{\raisebox{-1.5pt}{\includegraphics[height=1.05em]{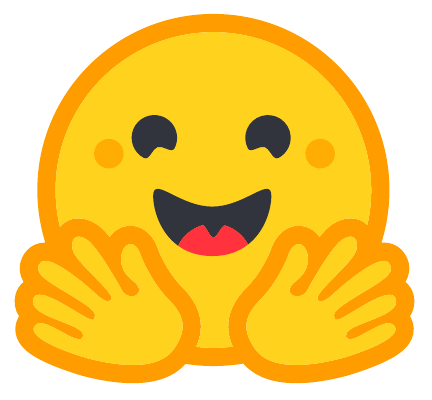}}\xspace}

\definecolor{greenbg}{RGB}{230, 255, 230}

\newtheorem{theorem}{Theorem}[section]

\newtheorem{corollary}[theorem]{Corollary}

\title{EntroPIC: Towards Stable Long-Term Training of LLMs via Entropy Stabilization with Proportional-Integral Control}

\author{
Kai Yang$^{1}$, 
Xin Xu$^{1,2}$, 
Yangkun Chen$^{1}$, 
Weijie Liu$^{1}$,\\
Jiafei Lyu$^{1}$,
Zichuan Lin$^{1}$, 
Deheng Ye$^{1}$,
Saiyong Yang$^{1,\dagger}$\\
\textbf{$^1$Tencent Hunyuan}\\
\textbf{$^2$The Hong Kong University of Science and Technology} \\ 
\Letter~\{kasperyang, stevesyang\}@tencent.com\\
}

\begin{document}
\maketitle

\renewcommand*{\thefootnote}{\fnsymbol{footnote}}
%\footnotetext{* Equal contribution.}
\footnotetext{$\dagger$ Corresponding Authors.}
\begin{abstract}
Long-term training of large language models (LLMs) requires maintaining stable exploration to prevent the model from collapsing into sub-optimal behaviors. Entropy is crucial in this context, as it controls exploration and helps avoid premature convergence to sub-optimal solutions. However, existing reinforcement learning methods struggle to maintain an appropriate level of entropy, as the training process involves a mix of positive and negative samples, each affecting entropy in different ways across steps. To address this, we propose \textbf{Entro}py stabilization via \textbf{P}roportional-\textbf{I}ntegral \textbf{C}ontrol (\textbf{EntroPIC}), a novel method that adaptively adjusts the influence of positive and negative samples by dynamically tuning their loss coefficients. This approach stabilizes entropy throughout training, ensuring efficient exploration and steady progress. We provide a comprehensive theoretical analysis for both on-policy and off-policy learning settings, demonstrating that EntroPIC is effective at controlling entropy in large-scale LLM training. Experimental results show that our method successfully maintains desired entropy levels, enabling stable and optimal RL training for LLMs.
\end{abstract}
% \let\thefootnote\relax\footnotetext{* First three authors contributed equally. ~~$^\dagger$ Corresponding Author: xxx.}

% \let\thefootnote\relax\footnotetext{* First two authors contributed equally. }
                % \let\thefootnote\relax\footnotetext{}

% \begin{center}
%     {\fontfamily{pcr}\selectfont
%         \begin{tabular}{rll}
%             \github & \textbf{Github Repo} & \href{https://github.com/XinXU-USTC/TFPI-temp.git}{[GitHub Page]} \\
%             \dataset & \textbf{Models} & \href{https://github.com/XinXU-USTC/TFPI-temp.git}{[Huggingface Models]} \\
%         \end{tabular}
%     }
% \end{center}

\begin{center}
    {\fontfamily{pcr}\selectfont
        \begin{tabular}{rll}
            \github & \textbf{Github Repo} & \href{https://github.com/yk7333/EntroPIC}{[GitHub Page]} \\
             \dataset & \textbf{Models} & \href{https://huggingface.co/yangkaiSIGS/EntroPIC-Nemotron-1.5b}{[Huggingface Models]} \\
        \end{tabular}
    }
\end{center}

\input{chapters/1_intro}

\input{chapters/2_related_work}

\input{chapters/3_preliminary}

\input{chapters/4_method}

\input{chapters/5_experiments}

\input{chapters/6_conclusion}
\clearpage

\bibliography{iclr2025_conference}
\bibliographystyle{iclr2025_conference}
\clearpage
\appendix
\clearpage
\input{appendix/A_proof}

\input{appendix/B_detail}

\input{appendix/C_Code}
\input{appendix/D_MoreResults}

\input{appendix/E_case}
% \input{appendix/experiment}
% \input{appendix/analysis}
% %\input{appendix/LLM_use}
% \input{appendix/results}

\end{document}

%% file: math_commands.tex
\usepackage{amsmath,amsfonts,bm}
\usepackage{multicol}
\def\eqref#1{equation~\ref{#1}}
\def\1{\bm{1}}

\DeclareMathAlphabet{\mathsfit}{\encodingdefault}{\sfdefault}{m}{sl}
\SetMathAlphabet{\mathsfit}{bold}{\encodingdefault}{\sfdefault}{bx}{n}

%% file: chapters/1_intro.tex
\begin{figure*}[h]
% \vspace{-0.4em}
    \centering
    \includegraphics[height=180pt, width=1.0\linewidth]{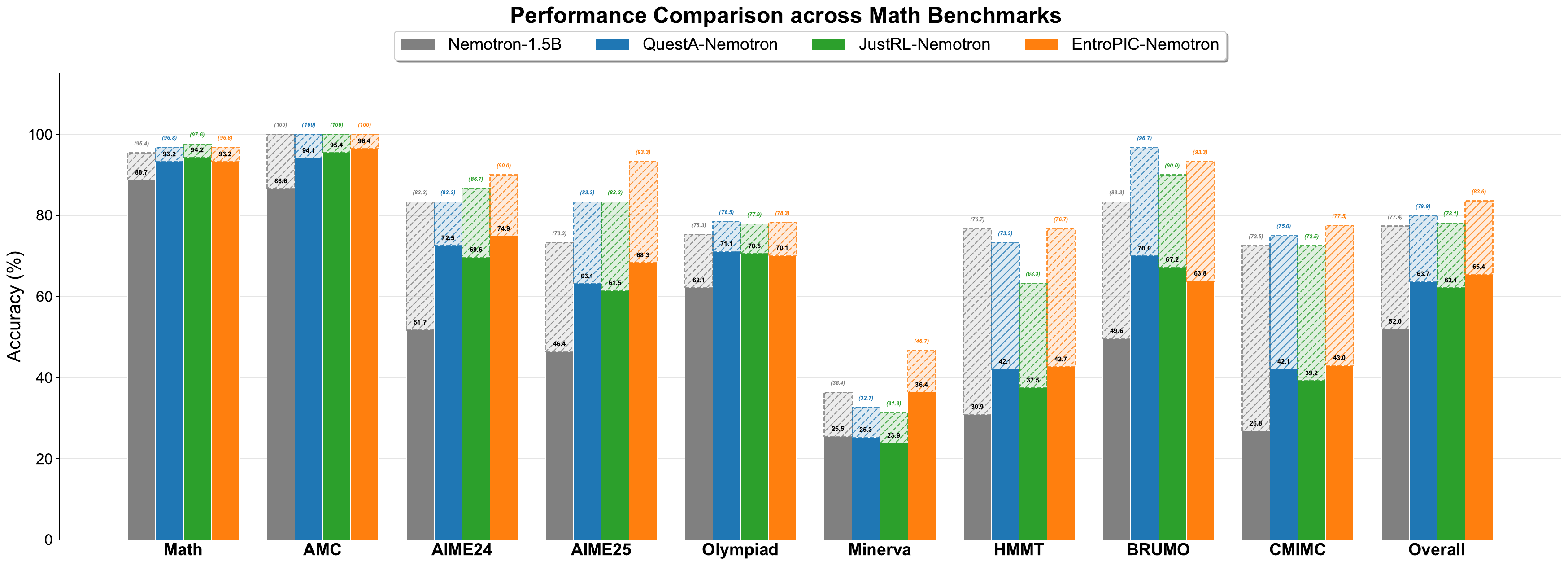}
    \vspace{-1em}
    \caption{Evaluation of 1.5B parameter models on a comprehensive suite of mathematical reasoning tasks. We utilize a diverse set of widely recognized benchmarks, including standard competitions like \textbf{AIME} and MathArena datasets such as \textbf{HMMT}. The solid regions represent average accuracy (\texttt{pass@1} or \texttt{avg@N}), while the upper dashed regions indicate the pass rate (\texttt{pass@N}). EntroPIC-Nemotron-1.5B demonstrates the highest overall performance compared to QuestA and JustRL methods, setting a new benchmark for 1.5B parameter models.}
    \label{entropic_reasoning}
    \vspace{-0.5em}
\end{figure*}

\section{Introduction}
Large language models (LLMs) have demonstrated exceptional performance across various natural language processing tasks, excelling in comprehension, generation, and reasoning ~\citep{achiam2023gpt, team2024gemini, liu2024deepseek}. In verifiable-reward domains, such as mathematics and programming, reinforcement learning with verifiable rewards (RLVR) ~\citep{guo2025deepseek} has proven highly effective. RLVR addresses the issue of reward hacking, common in reward models, by directly comparing model outputs to ground-truth answers. In this context, a reinforcement learning algorithm capable of sustaining long-term training is essential for continuous model improvement. Notably, output diversity, quantified by entropy, plays a critical role in both the effectiveness and stability of the training process \citep{cui2025entropy}.

Entropy serves as a key measure in reinforcement learning (RL), quantifying agent exploration and preventing premature convergence to local optima. Maintaining an appropriate level of entropy is vital: too low entropy causes premature convergence ~\citep{arora2023stable, li2025entropy, zawistowski2024unused}, while too high entropy leads to noisy outputs and instability. Foundational RL research has successfully integrated entropy into optimization algorithms to balance exploration and exploitation in classical environments ~\citep{haarnoja2018soft, zhao2019maximum, eysenbach2021maximum, shi2019soft}. However, RL application to LLMs is challenged by the fact that initial models are pre-trained and start with low entropy. Low regularization coefficients fail to preserve exploration, while high coefficients degrade performance and increase sensitivity to hyperparameters ~\citep{he2025skywork, cui2025entropy}.

Current entropy regulation methods can be divided into two main approaches: (1) masking or down-weighting tokens to enhance diversity ~\citep{cui2025entropy, du2025ulorl, zhu2025surprising}, and (2) embedding entropy bonuses into the reward or loss functions ~\citep{zhang2024entropy, li2025entropy, he2025skywork}. While these methods slow the decline in entropy, they struggle to maintain entropy near a desired target without sacrificing model performance. Mask-based techniques lose token gradients, while weight and bonus methods suffer from sensitivity to hyperparameters and added complexity, hindering scalability ~\citep{he2025skywork}. Furthermore, most methods focus on off-policy settings, even though on-policy training has demonstrated higher performance potential. However, early-stage uncontrollable entropy fluctuations during large dataset training remain a challenge, and as a result, on-policy entropy control has been less explored.

Empirical research shows that positive-sample training reduces entropy (negative samples increase it) and that policy entropy is influenced by the covariance between action log-probabilities and the advantage function ~\citep{zhu2025surprising, cui2025entropy}. This insight frames entropy regulation as a balance between these opposing effects. In this work, we first prove theoretically that positive samples decrease entropy (negative samples increase it) and then introduce EntroPIC. EntroPIC uses Proportional (P) and Proportional-Integral (PI) control to dynamically adjust the weight of positive and negative samples. By using target entropy and tracking both historical and current deviations, it stabilizes training entropy, offering convergence guarantees.

As shown in Figure \ref{entropic_plt}, EntroPIC stabilizes entropy and gradient updates by adjusting the weight of high-probability tokens. Dynamic tuning of coefficients enables targeted entropy control without requiring global adjustments. This method stabilizes entropy for both on-policy and off-policy training, and its effectiveness has been validated through successful training on over 1M prompts, demonstrating its suitability for large-scale LLM training in industrial settings.

%% file: chapters/2_related_work.tex
\begin{figure*}[h]
% \vspace{-0.2em}
\centering
\includegraphics[width=\linewidth]{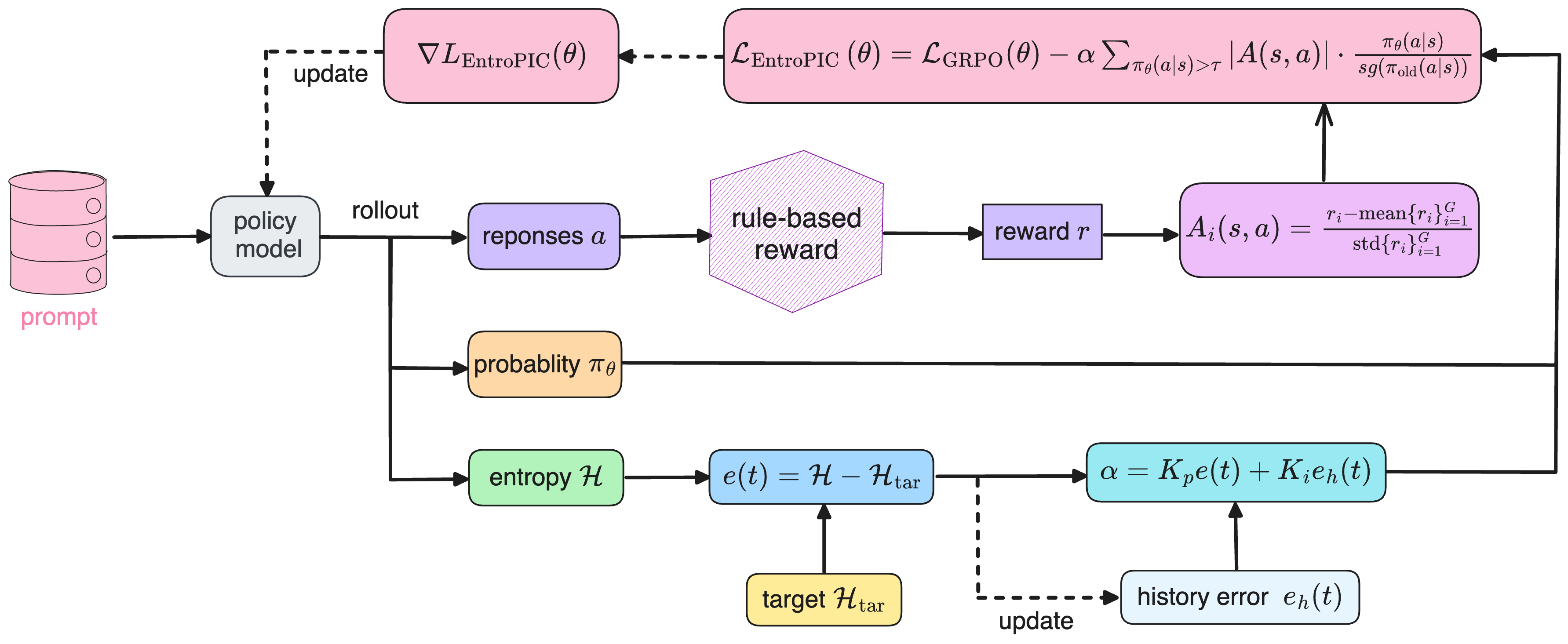}
\caption{Overview of EntroPIC. The method calculates a correction factor based on historical and current entropy values relative to the target entropy and adjusts the weights of high-probability positive and negative samples to achieve entropy control.}
\label{entropic_plt}
\vspace{-0.5em}
\end{figure*}
\section{Related Work}
\textbf{Reinforcement Learning with Large Language Models.}
Recent advances in aligning large language models (LLMs) with human preferences have heavily relied on reinforcement learning techniques, particularly Reinforcement Learning from Human Feedback (RLHF) ~\citep{achiam2023gpt, ouyang2022training, korbak2023pretraining, hurst2024gpt}. These methods typically involve training a reward model based on human or AI feedback, followed by policy optimization using algorithms like PPO ~\citep{schulman2017proximal} and GRPO ~\citep{shao2024deepseekmath}. Recently, RLVR has made significant progress in fields like mathematics, coding, science, and complex instructions. RLVR eliminates the need for a reward model and is nearly immune to overfitting, offering a breakthrough in performance ~\citep{lambert2024tulu, o1, zhang2025merf, guo2025deepseek, team2025kimi, zhang2025supervised}. With RLVR, the performance ceiling is no longer limited by the quality of the reward model but instead depends largely on the control of entropy ~\citep{cui2025entropy, yu2025dapo}. Our research demonstrates that maintaining an appropriate entropy level enhances model diversity and boosts its performance potential.

\textbf{Entropy in Policy Optimization.}
In reinforcement learning, exploration is essential for achieving the upper limit of model performance ~\citep{sutton1998reinforcement}. Many works related to exploration have achieved success in gaming and robotics scenarios ~\citep{burda2018exploration, yang2024exploration, lobel2023flipping}. One major category of methods focuses on enhancing exploration by increasing entropy. Entropy regularization has long been recognized as a critical component in policy gradient methods to promote exploration ~\citep{williams1991function, mnih2016asynchronous, haarnoja2017reinforcement, kim2023accelerating}. Maintaining entropy at an appropriate level prevents premature convergence to deterministic and potentially suboptimal policies ~\citep{haarnoja2018soft, schulman2017proximal}. Recent studies have further investigated entropy dynamics in the context of LLM alignment, proposing methods such as modifying entropy bonuses ~\citep{zhang2024entropy, li2025entropy, tan2025gtpo}, importance sampling weights ~\citep{vanlioglu2025entropy, yin2024entropy, cui2025entropy, su2025gppo}, or entropy regularization terms ~\citep{adamczyk2025average, he2025skywork}. While these methods can slow the decline in entropy, they still struggle to maintain entropy at a high, stable level during long-term training.

%% file: chapters/3_preliminary.tex
\section{Preliminary}
\textbf{Reinforcement Learning with Verifiable Rewards (RLVR).}  
RLVR is a framework for training LLMs on tasks with objectively verifiable outcomes, such as mathematics and programming. Unlike RLHF, RLVR directly computes rewards by comparing model outputs to ground-truth solutions, eliminating the need for reward models and mitigating reward hacking. For a given prompt $x$, an LLM policy $\pi_\theta$ generates a token sequence $y = \{y_1, \cdots, y_T\}$. The goal is to maximize the expected reward:
\[
\max_\theta J(\theta) := \mathbb{E}_{x \sim D, y \sim \pi_\theta(x)} [r(y)],
\]  
where $D$ is the training distribution and $r(y)$ is the reward for output $y$. To optimize this objective, policy gradient methods estimate the gradient of $J(\theta)$ as:
\[
\nabla_\theta J(\theta) = \mathbb{E}_{x \sim D, y \sim \pi_\theta(x)} \left[ \sum_{t=1}^T \nabla_\theta \log \pi_\theta(y_t|y_{<t}) A_t \right].
\]  

Here, $A_t$ is the advantage at step $t$, reflecting the value of selecting token $y_t$ given the context $y_{<t}$.  

In Group Relative Policy Optimization (GRPO) ~\citep{shao2024deepseekmath}, a representative on-policy method, advantages are normalized within groups of sampled responses to reduce variance. For each prompt $x$, $K$ responses $\{y^1, \cdots, y^K\}$ are sampled, and the advantage for response $y^k$ is:
\[
A^k = \frac{r(y^k) - \text{mean}(r(y^{1:K}))}{\text{std}(r(y^{1:K})) + \epsilon},
\]  
where $\epsilon$ is a small constant. This group-normalized advantage is used for gradient updates, ensuring robustness to reward scale variations.

For constrained policy updates, Proximal Policy Optimization (PPO) ~\citep{schulman2017proximal} uses a clipped surrogate objective:
{\small
$$
L(\theta) = -\mathbb{E}_{\pi_{\theta_\text{old}}} \left[\min\left(r_t(\theta) A_t, \; \text{clip}(r_t(\theta), 1-\epsilon, 1+\epsilon) A_t\right)\right]
$$ 
}
,where $r_t(\theta) = \frac{\pi_\theta(y_t|y_{<t})}{\pi_{\theta_{\text{old}}}(y_t|y_{<t})}$ is the advantage estimate at time step $t$.

This clipping term limits excessive policy updates, balancing optimization progress with training stability.

\textbf{Policy Entropy in Large Language Models.}  
Policy entropy measures the uncertainty of token selection in LLM generation, balancing exploration and exploitation. For a policy $\pi_\theta$ on training data $D$, the average token-level entropy is defined as:
$$
\mathcal{H}(\pi_\theta, D) = -\mathbb{E}_{x \sim D, y \sim \pi_\theta(x)} \left[ \frac{1}{|y|} \sum_{t=1}^{|y|} \log \pi_\theta(y_t|y_{<t}) \right].
$$  

This quantifies the randomness of token predictions. Maintaining appropriate entropy is critical: low entropy causes mode collapse (deterministic, suboptimal outputs), while high entropy leads to unstable training and noisy generation.

Notably, LLM-RL starts with pretrained policies with low entropy, making fixed-coefficient entropy regularization ineffective due to hyperparameter sensitivity. Our analysis shows that entropy dynamics are influenced by the covariance between $\log \pi_\theta(y_t|y_{<t})$ and the advantage $A_t$: positive samples (high $A_t$) reduce entropy, while negative samples (low $A_t$) increase it.

\textbf{Proportional-Integral Control.}  
PI control is a feedback mechanism that adjusts control inputs based on error signals to track a target setpoint. In continuous time, the control law is:
\[
u(t) = K_p \cdot e(t) + K_i \cdot \int_0^t e(\tau) d\tau,
\]  
where $e(t) = r(t) - y(t)$ is the error between the target setpoint $r(t)$ and the measured variable $y(t)$, with $K_p$ and $K_i$ being the proportional and integral gains, respectively. The proportional term provides an immediate response to current errors, while the integral term accumulates historical errors to eliminate steady-state bias.

In discrete-time training (e.g., LLM-RL), the integral term is approximated by a summation. For the $n$-th step, the discrete PI control law becomes:
\[
u(n) = K_p \cdot e(n) + K_i \cdot \sum_{k=1}^{n-1} e(k),
\]  
where $e(n) = r - y(n)$ is the error in step $n$, and the summation replaces the integral.  

In our work, PI control is adapted to entropy regulation: the target setpoint $r$ is the desired entropy, $y(n)$ is the measured entropy at step $n$, and $e(n)$ dynamically adjusts the sample weight coefficients.

%% file: chapters/4_method.tex
\section{Adjust Entropy via PI Control}
\subsection{The Impact of Positive and Negative Samples on Entropy}
To effectively control entropy during RL training, it is essential to understand the impact of positive and negative samples on training entropy. The NSR method~\citep{zhu2025surprising} experimentally demonstrates that positive samples tend to decrease entropy, while negative samples tend to increase it. We first provide a theoretical explanation for these observations, showing that using only positive or negative samples during training with the RLVR binary reward distribution results in a decrease or increase in entropy, respectively.

\begin{corollary}\label{corollary1}
When the reward values for positive and negative samples follow a binary distribution, and the expected advantage equals zero, training exclusively with positive/negative samples using a policy gradient method will result in a decrease/increase of the policy's entropy.
\end{corollary}

The complete proof is provided in Appendix \ref{proof_co}. This corollary illustrates that the use of positive samples results in a decrease in entropy, while negative samples cause an increase. Based on these findings, we will modify the loss weights for positive and negative samples to dynamically regulate entropy during training.

\subsection{Adjusting Weights to Control Entropy}
\label{entropic_method}
Building on the previous discussion, training with a mixture of positive and negative samples can lead to an imbalance, resulting in either excessively high or low entropy. The policy gradient loss function for sequence modeling is defined as follows:
\begin{equation*}  
L(\theta) = -\mathbb{E}_{a \sim \pi_\theta}[A(s,a) \log \pi_\theta(a|s)].
\end{equation*}

To stabilize entropy, we propose the EntroPIC method, which uses PI control to dynamically adjust the loss coefficients for positive and negative samples based on the discrepancy between the current entropy value and the target entropy. The modified loss function is:
{\small
\begin{align*}
L(\theta) &= -\mathbb{E}_{a \sim \pi_\theta}\Big[\mathbb{I}(A(s,a) > 0) (1+\alpha) A(s,a) \log \pi_\theta(a|s) \\
&+ \mathbb{I}(A(s,a) < 0) (1-\alpha) A(s,a) \log \pi_\theta(a|s)\Big],
\end{align*}
}
where $\mathbb{I}$ is the indicator function and $\alpha$ is a balancing coefficient that adjusts the influence of positive and negative samples. $\alpha$ is determined by the PI control formula:
\begin{align*}
\alpha^t = K_p(\mathcal{H}^t - \mathcal{H}^t_\text{tar}) + K_i \sum_{k=1}^{t-1} (\mathcal{H}^k - \mathcal{H}^k_\text{tar}).
\end{align*}

Here, $\mathcal{H}^t$ is the entropy at time step $t$, $\mathcal{H}^t_\text{tar}$ is the target entropy, and $K_p$ and $K_i$ are the proportional and integral control coefficients. When $\alpha = 0$, the loss function is equivalent to the original policy gradient loss. If $\alpha = 1$, the loss focuses only on positive samples, reducing entropy; if $\alpha = -1$, only negative samples are updated, increasing entropy. This mechanism enables dynamic regulation of policy entropy during training.

\begin{theorem}\label{theorem1}
When training with an on-policy method and using P-control or PI-control (i.e., when $K_p > 0$ and $K_i \geq 0$), the error $e_t$ between the policy's entropy and the target entropy $\mathcal{H}^t_\text{tar}$ converges to zero:  
\[
\lim_{k \to \infty} e_k \to 0.
\]
\end{theorem}

The proof can be found in Appendix \ref{proof_theorem1}. This theorem shows that using P-control or PI-control ensures reliable entropy regulation toward the target entropy for on-policy methods.

For off-policy training, the situation changes as the sampling and training policies are different. The policy gradient loss is:
\begin{align*}  
L(\theta) &= -\mathbb{E}_{a \sim \mu(\cdot|s)} [ \min \left( \rho(a|s) \cdot A(s,a), \right. \\
&\quad\quad\quad\quad \left. \text{clip}(\rho(a|s), 1-\epsilon_\text{low}, 1 + \epsilon_\text{high}) A(s,a) \right) ],
\end{align*}
where $\rho(a|s)=\frac{\pi_\theta(a|s)}{\mu(a|s)}$ is the importance sampling ratio. Our entropy control method also adjusts the sample weights, resulting in the modified loss:
\begin{align*}  
L(\theta) &= -\mathbb{E}_{a \sim \mu} [ \mathbb{I}(A(s,a) > 0) (1+\alpha)\rho(a|s) \cdot A(s,a) + \\ &\quad\quad\quad\quad\quad \mathbb{I}(A(s,a) < 0) (1-\alpha)\rho(a|s) \cdot A(s,a) ].
\end{align*}

\begin{theorem}\label{theorem2}
When training with an off-policy method using P-control with the above loss (i.e., $K_p > 0$ and $K_i = 0$), a steady-state error in entropy will occur. Only PI-control (i.e., $K_p > 0$ and $K_i > 0$) ensures that the error $e_t$ between the policy's entropy and the target entropy $\mathcal{H}^t_\text{tar}$ converges to zero:
\[
\lim_{k \to \infty} e_k \to 0.
\]
\end{theorem}

The proof is found in Appendix \ref{proof_theorem2}. From Theorems \ref{theorem1} and \ref{theorem2}, we conclude that for on-policy training, P-control is sufficient to achieve convergence to the target entropy, while for off-policy training, PI-control is necessary. Our method is versatile, with no restrictions on the initial or target entropy, and can adjust entropy in both directions, which most existing methods cannot. This approach guarantees convergence to the target entropy.

\subsection{Simplify Loss by Controlling High-Probability Samples}
\label{sec:highprob}
In Section \ref{entropic_method}, we introduced entropy control by adjusting the weights of all positive and negative samples. However, modifying low-probability samples can lead to substantial entropy changes and affect stability~\citep{cui2025entropy}. We now explore whether we can achieve similar entropy control by adjusting the weights of only a small subset of tokens, specifically high-probability samples.

\begin{figure*}[h]
    \vspace{-0.5em}
    \centering
    \subfigure{
    \begin{minipage}[t]{0.4\linewidth}
    \centering
    \includegraphics[width=\linewidth]{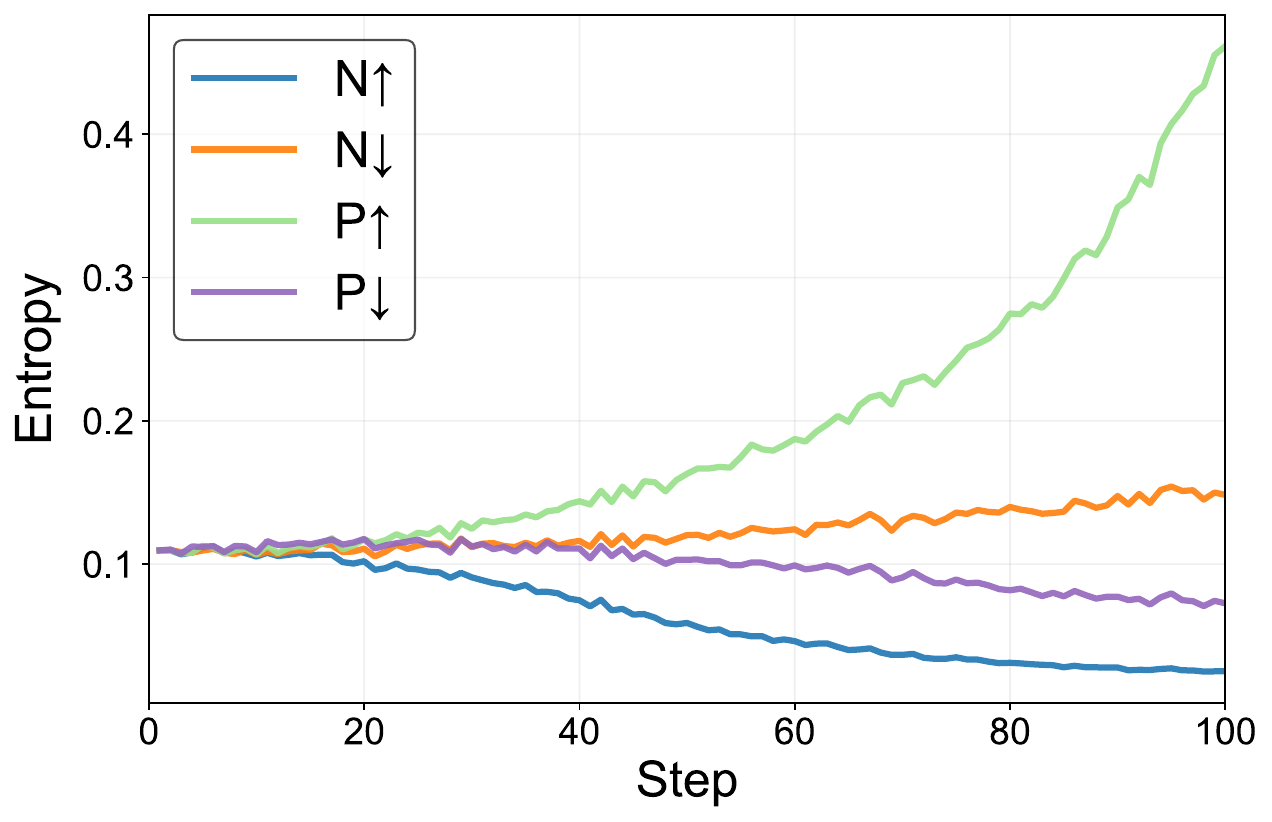}
    \end{minipage}
    }
    \subfigure{
    \begin{minipage}[t]{0.48\linewidth}
    \centering
    \includegraphics[width=\linewidth]{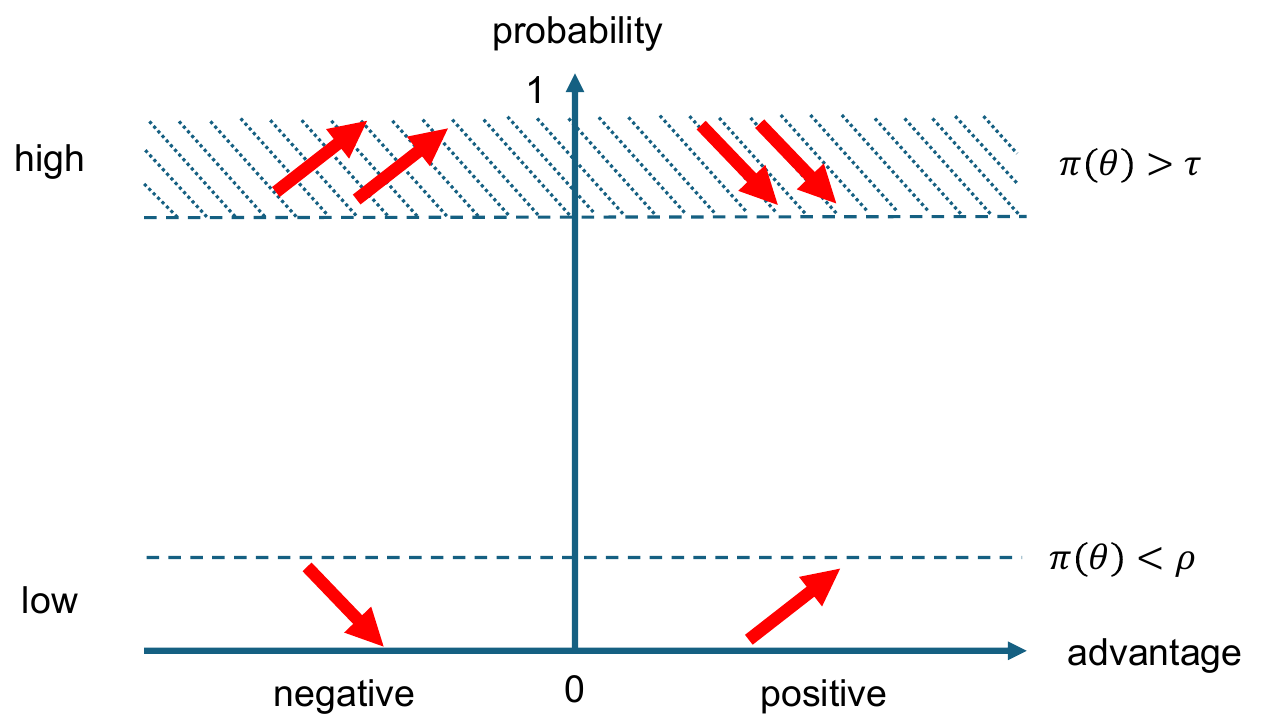}
    \end{minipage}
    }
    \caption{Investigating how masking positive/negative and high/low-probability tokens affects entropy during training. Left: Entropy variation after masking different token groups (P↑: high-prob. positive, P↓: low-prob. positive, N↑: high-prob. negative, N↓: low-prob. negative). Right: Comparison of entropy changes between high- and low-probability tokens.}
    \label{entropy_mask}
    % \vspace{-0.5em}
\end{figure*}

\begin{corollary}\label{corollary2}
When the control coefficient $\alpha$ is applied only to high-probability samples, i.e., tokens satisfying $\pi_\theta(y_t|y_{<t}) > \tau$ for a fixed threshold $\tau$,
the convergence properties of Theorems~\ref{theorem1} and~\ref{theorem2} still hold under the same conditions.
\end{corollary}

The proof can be viewed as Appendix \ref{proof_corollary}. As shown in Appendix \ref{proof_theorem1}, entropy changes are positively correlated with token probabilities. Therefore, controlling the weights of high-probability tokens should yield similar results with minimal disruption to the original training process. To test this hypothesis, we conducted ablation experiments where we masked high-probability and low-probability positive and negative samples, observing entropy changes during training, as shown in Figure \ref{entropy_mask}. The results confirm that modifying the weights of high-probability samples produces more significant entropy changes than adjusting low-probability tokens.

\begin{figure*}[!h]
    \centering
    \includegraphics[width=0.8\linewidth]{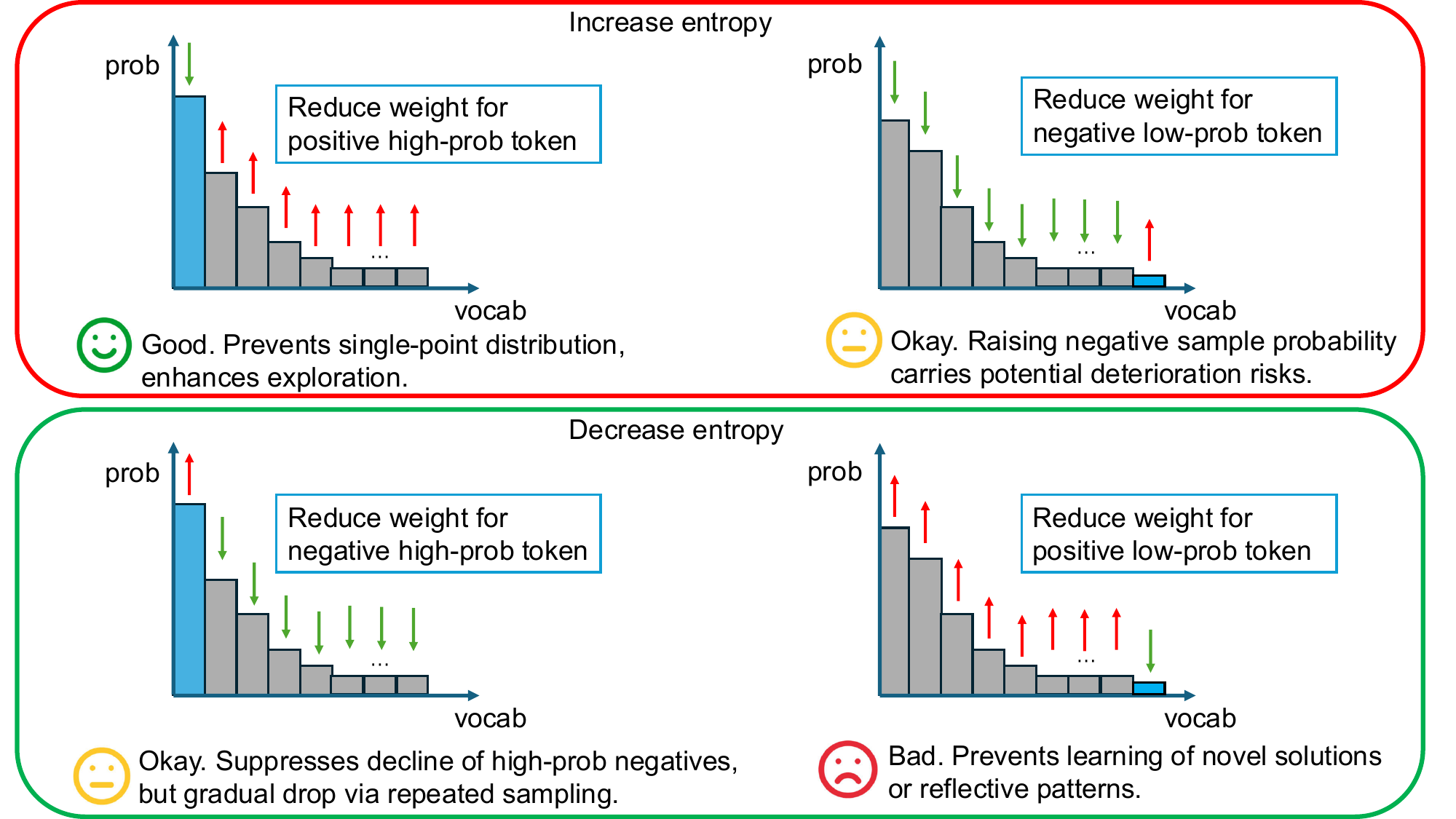}
    \caption{Schematic of entropy control via weight modulation. Modulating weights for high-probability tokens results in better performance than for low-probability ones.}
    \label{highprob}
    \vspace{-0.5em}
\end{figure*}

Modifying only high-probability tokens offers several advantages over adjusting low-probability ones:  
1) \textbf{Easier Identification:} High-probability tokens are easily identified—tokens with probabilities greater than 95\% are considered high-probability. Low-probability tokens are more challenging to define due to dependencies on sampling strategies like temperature, top-p, and top-k.  
2) \textbf{Enhanced Exploration:} Reducing the weight of high-probability positive samples encourages exploration (Figure \ref{highprob}), while reducing the weight of low-probability negative samples can avoid suppressing rare events, which could degrade model performance.

In conclusion, EntroPIC focuses on adjusting the weights of high-probability tokens to achieve optimal entropy control. The simplified loss function, which uses only high-probability tokens for regulation, is
\begin{align*}
\mathcal{L}(\theta) &= -(1+\alpha) \sum_{\substack{\pi_\theta(a|s)>\tau ,\\  A(s,a)>0}} A(s,a) \log\pi_\theta(a|s)  - (1-\alpha) \sum_{\substack{\pi_\theta(a|s)>\tau ,\\ A(s,a)<0}} A(s,a) \log \pi_\theta(a|s)  \\& \quad\quad\quad- \sum_{\pi_\theta(a|s)<\tau} A(s,a) \log \pi_\theta(a|s) \\
&= \mathcal{L}_\text{pg}(\theta) - \alpha \sum_{\pi_\theta(a|s)>\tau} |A(s,a)| \log\pi_\theta(a|s),
\end{align*}
where $\mathcal{L}_\text{pg}(\theta) = -\sum A(s,a)\log \pi_\theta(a|s)$ denotes the policy gradient loss. For off-policy PPO-like losses, the entropy-controlled loss becomes:
\begin{equation*}
\mathcal{L}(\theta) = \mathcal{L}_\text{pg}(\theta) - \alpha \sum_{\pi_\theta(a|s)>\tau} |A(s,a)| \frac{\pi_\theta(a|s)}{\mu(a|s)}.
\end{equation*}
In both on-policy and off-policy scenarios, the final modified loss is:
\begin{equation*}
\mathcal{L}(\theta) = \mathcal{L}_\text{pg}(\theta) - \alpha \sum_{\pi_\theta(a|s)>\tau} |A(s,a)| \frac{\pi_\theta(a|s)}{sg(\pi(a|s))},
\end{equation*}
where $sg$ denotes the stop-gradient operation, and $\pi(a|s)$ is the sampling policy. Although this loss differs slightly from the on-policy version, their gradients are identical.

%% file: chapters/5_experiments.tex
\section{Experiment}
\label{exp}

\subsection{Settings}
We evaluate the performance of non-thinking models using Qwen3-8B-Base \citep{yang2025qwen3} on mathematical tasks to validate our method. Initially, the model is fine-tuned with supervised fine-tuning to endow it with basic mathematical capabilities. For RL training, we use several datasets, including DAPO-MATH-17K \citep{yu2025dapo}, OpenReasonerZero \citep{hu2025open}, and DeepScaleR \citep{deepscaler2025}. During training, evaluation is conducted on OMNI-MATH \citep{gao2024omnimathuniversalolympiadlevel}, AIME2024, AIME2025, AMC \citep{lightman2023let}, MATH \citep{lightman2023let}, and OlympiadBench \citep{he2024olympiadbench}, and the average accuracy across these datasets is reported. We adopt GRPO as the baseline RL algorithm and compare EntroPIC with other entropy control methods, including Clip\_cov, KL\_cov \citep{cui2025entropy}, DMMPT \citep{du2025ulorl}, NSR \citep{zhu2025surprising}, and AEC \citep{he2025skywork}.

In addition, we conduct experiments on reasoning models to evaluate the effectiveness of EntroPIC under explicit chain-of-thought generation. We adopt OpenReasoning-Nemotron-1.5B \citep{moshkov2025aimo2} as the base model and train it on the Polaris dataset \citep{an2025polaris}. For evaluation, beyond the standard mathematical benchmarks, we additionally include three MathArena evaluation datasets—HMMT, BRUMO, and CMIMC  \citep{balunovic2025matharena}—to comprehensively assess the model’s reasoning capabilities across diverse reasoning-intensive tasks. We compare EntroPIC with state-of-the-art models trained on Nemotron-1.5B, namely QuestA \citep{li2025questa} and JustRL \citep{he2025justrl}, to evaluate its effectiveness on reasoning models.

For entropy control, we apply the default PI-control setup with coefficients $K_p = 1$ , $K_i = 0.01$ and high-probability threshold $\tau=0.95$. In each rollout step, 8 responses per prompt are sampled for a batch of 512 prompts using a temperature of 0.6. The target entropy for EntroPIC is set to 0.1, while the configurations for other methods adhere to the default settings in their respective papers. More details can be found in Appendix \ref{appendix_setting}.

\subsection{Results}

\subsubsection{Entropy Control Experiments}

To verify that the proposed method is consistent with the theoretical analysis and whether it can control the entropy at the target value, we performed a simple entropy control experiment to evaluate the EntroPIC method. We examine both on-policy and off-policy reinforcement learning setups to validate whether the PI control mechanism can effectively stabilize entropy throughout training.

\begin{figure*}[h]
    \centering
    % 第一张图的内容 (Left)
    \subfigure{
    \begin{minipage}[t]{0.23\linewidth} % 调整宽度为 0.24 以容纳4张图
    \centering
    \includegraphics[width=\linewidth]{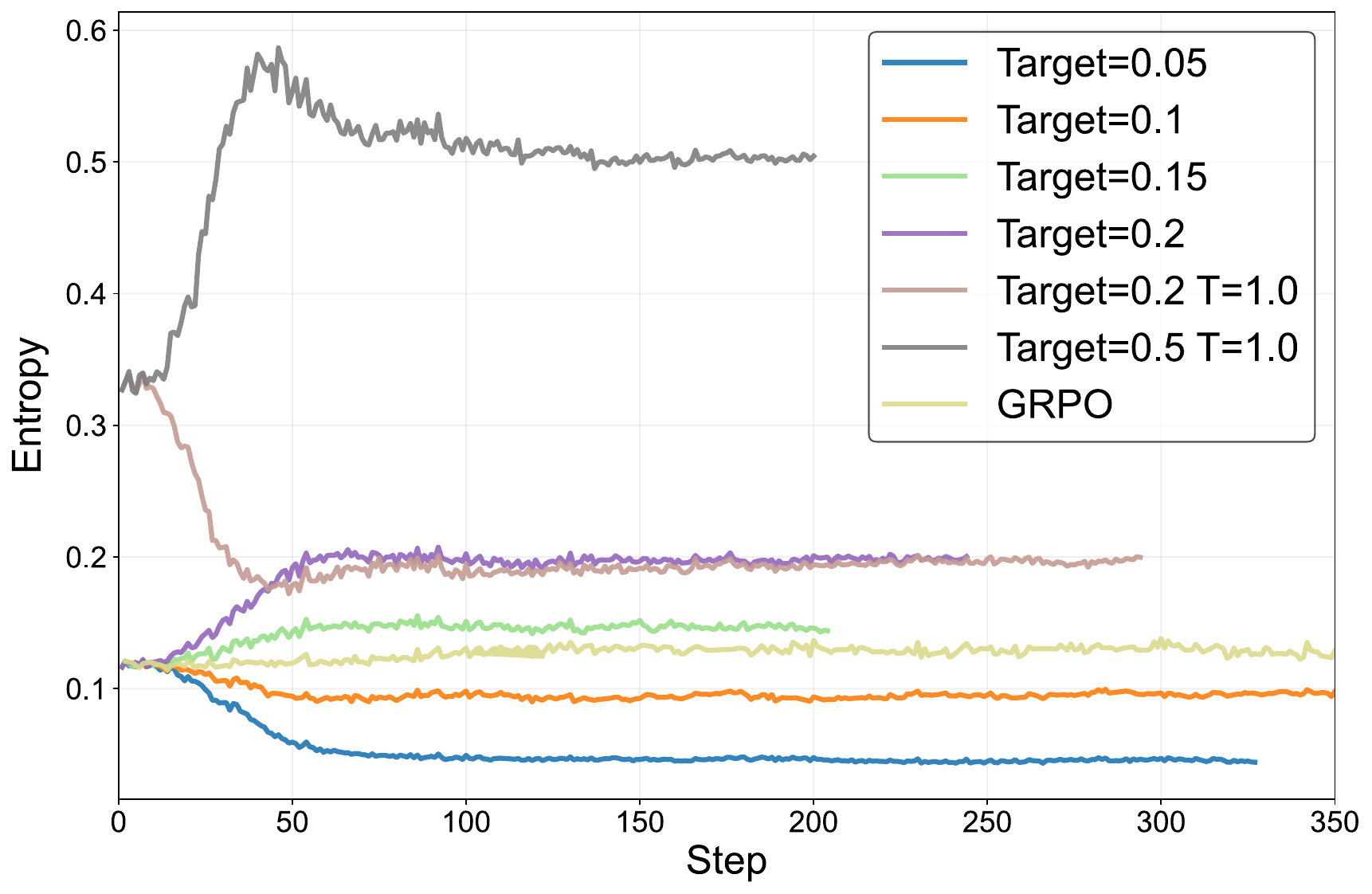}
    % \caption{Optional sub-caption} % 如果需要子标题可在此添加
    \end{minipage}
    }
    % 第一张图的内容 (Right)
    \subfigure{
    \begin{minipage}[t]{0.23\linewidth}
    \centering
    \includegraphics[width=\linewidth]{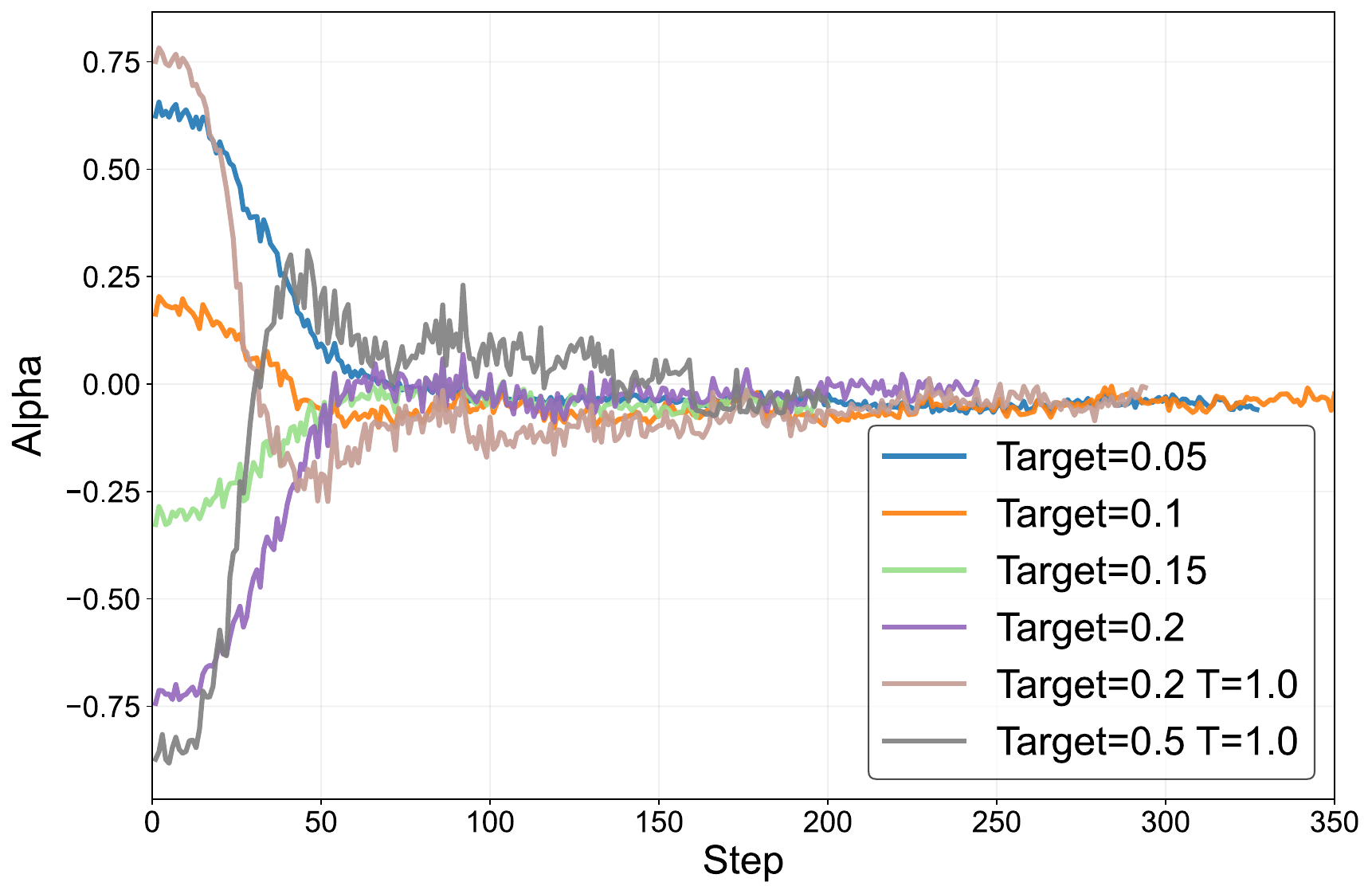}
    \end{minipage}
    }
    % 第二张图的内容 (Left)
    \subfigure{
    \begin{minipage}[t]{0.23\linewidth}
    \centering
    \includegraphics[width=\linewidth]{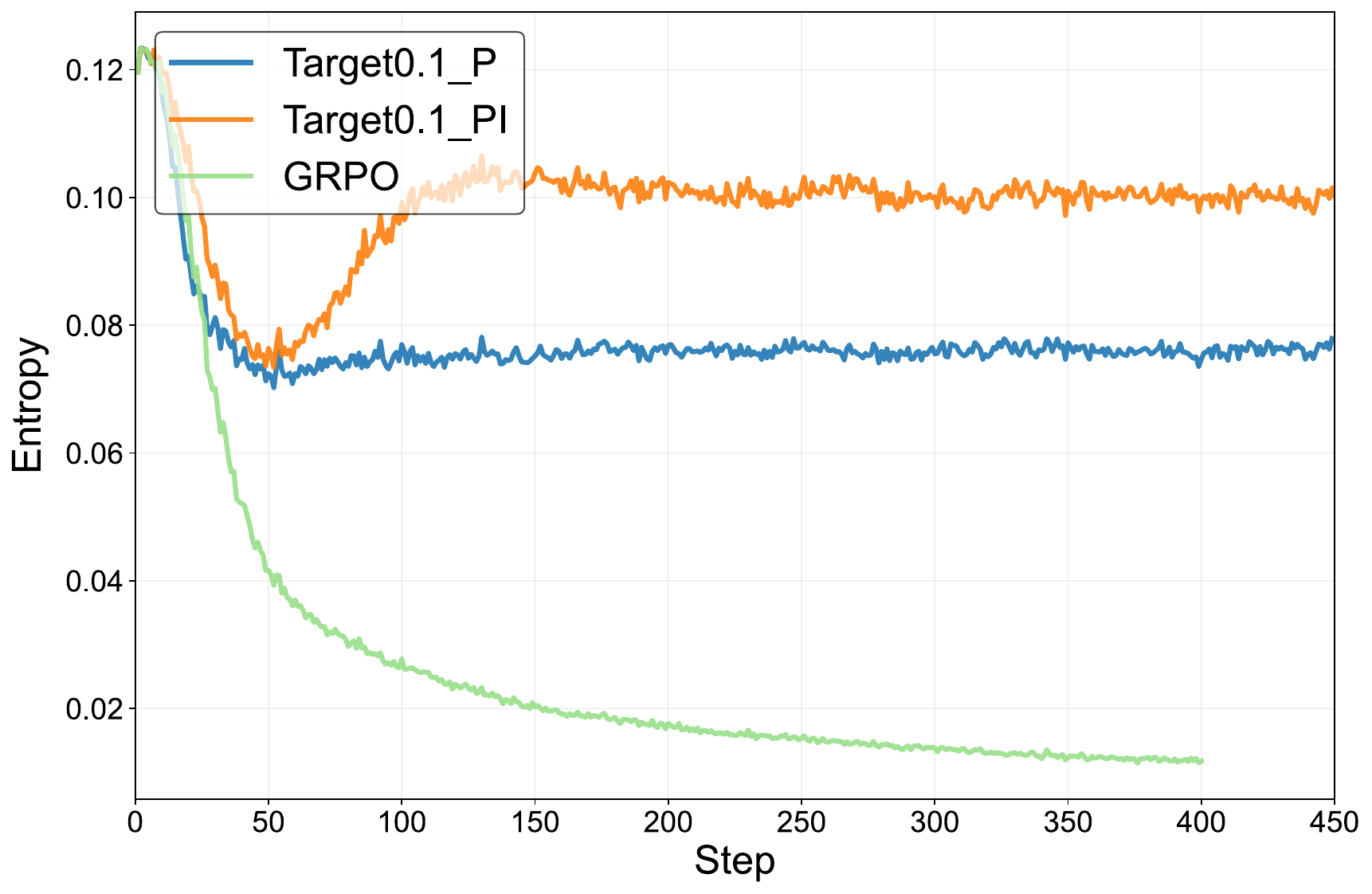}
    \end{minipage}
    }
    % 第二张图的内容 (Right)
    \subfigure{
    \begin{minipage}[t]{0.23\linewidth}
    \centering
    \includegraphics[width=\linewidth]{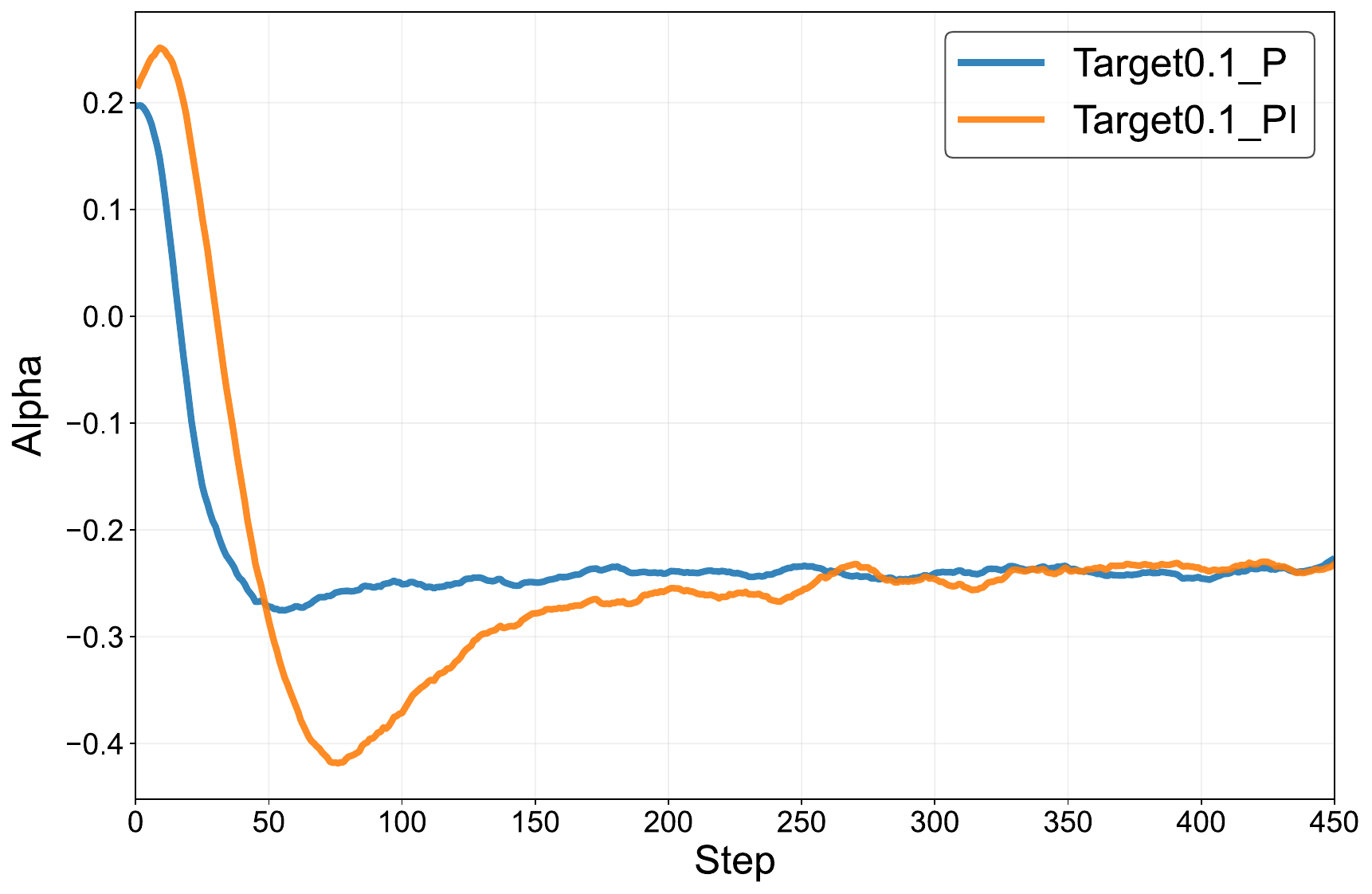}
    \end{minipage}
    }
    % \vspace{-1em}
    \caption{Results on entropy control experiments. \textbf{Left two panels:} On-policy results showing the entropy stabilization process and variation of adaptive coefficient $\alpha$. \textbf{Right two panels:} Off-policy results showing the entropy curve and control response, demonstrating that only PI control successfully stabilizes entropy at the target value.}
    \label{entropy}
    % \vspace{-0.5em}
\end{figure*}

As shown in Figure~\ref{entropy}, EntroPIC successfully stabilizes entropy near the desired target in both on-policy training and off-policy training. The left plot depicts the convergence of entropy over time, while the second plot shows how the adaptive coefficient $\alpha$ dynamically adjusts to maintain stability. In contrast, training without entropy control exhibits continual entropy decay. Similarly, in off-policy settings, we observe that only the full PI control can maintain entropy around the target value, whereas the P-only control fails to achieve stable regulation. These findings empirically confirm our theoretical claim that positive and negative sample weighting, governed by PI feedback, effectively controls entropy dynamics.

\subsubsection{Large-scale Training Results}
\begin{figure*}[h]
    \centering
    \subfigure{
    \begin{minipage}[t]{0.305\linewidth}
    \centering
    \includegraphics[width=\linewidth]{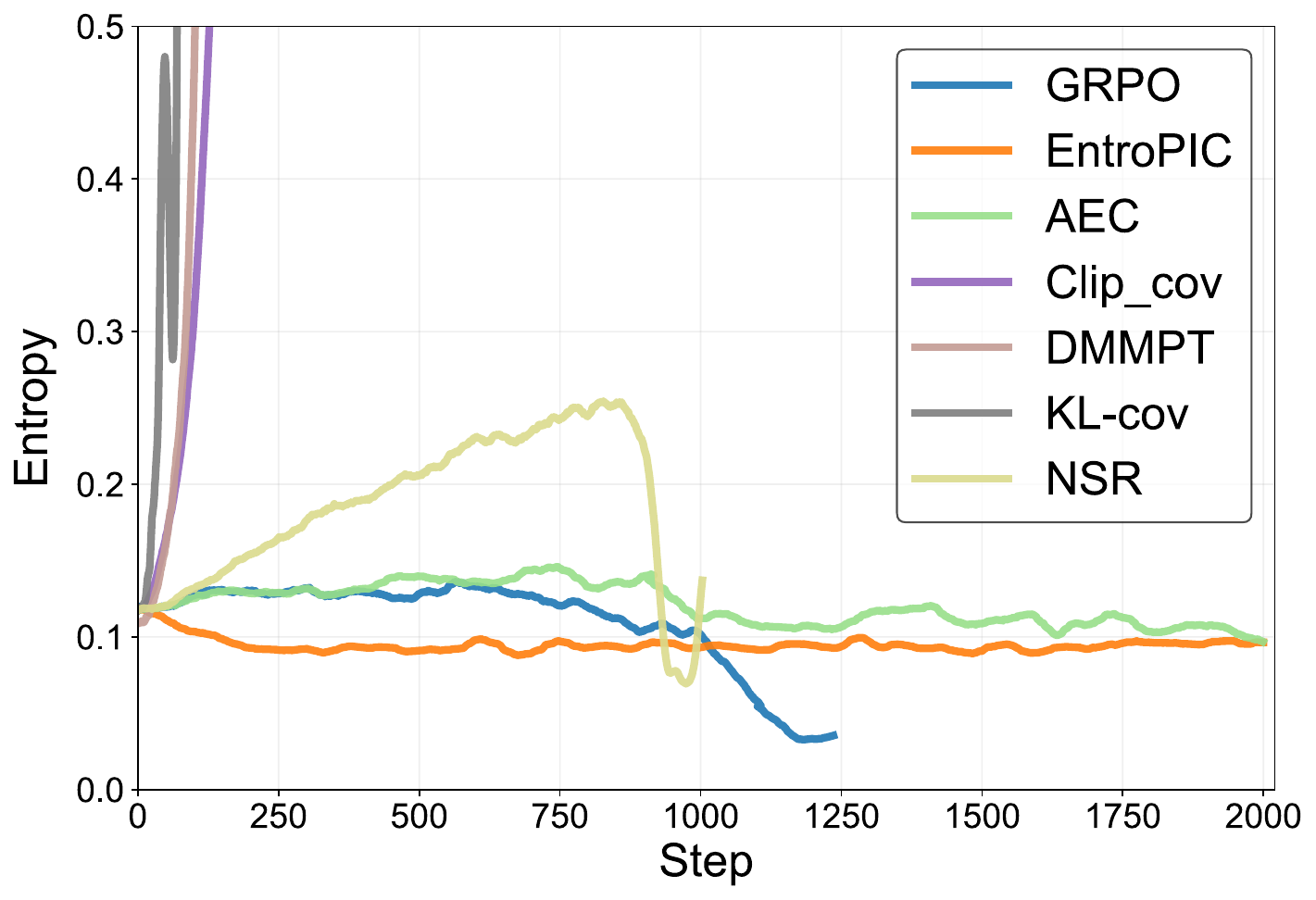}
    \end{minipage}
    }
    \subfigure{
    \begin{minipage}[t]{0.3\linewidth}
    \centering
    \includegraphics[width=\linewidth]{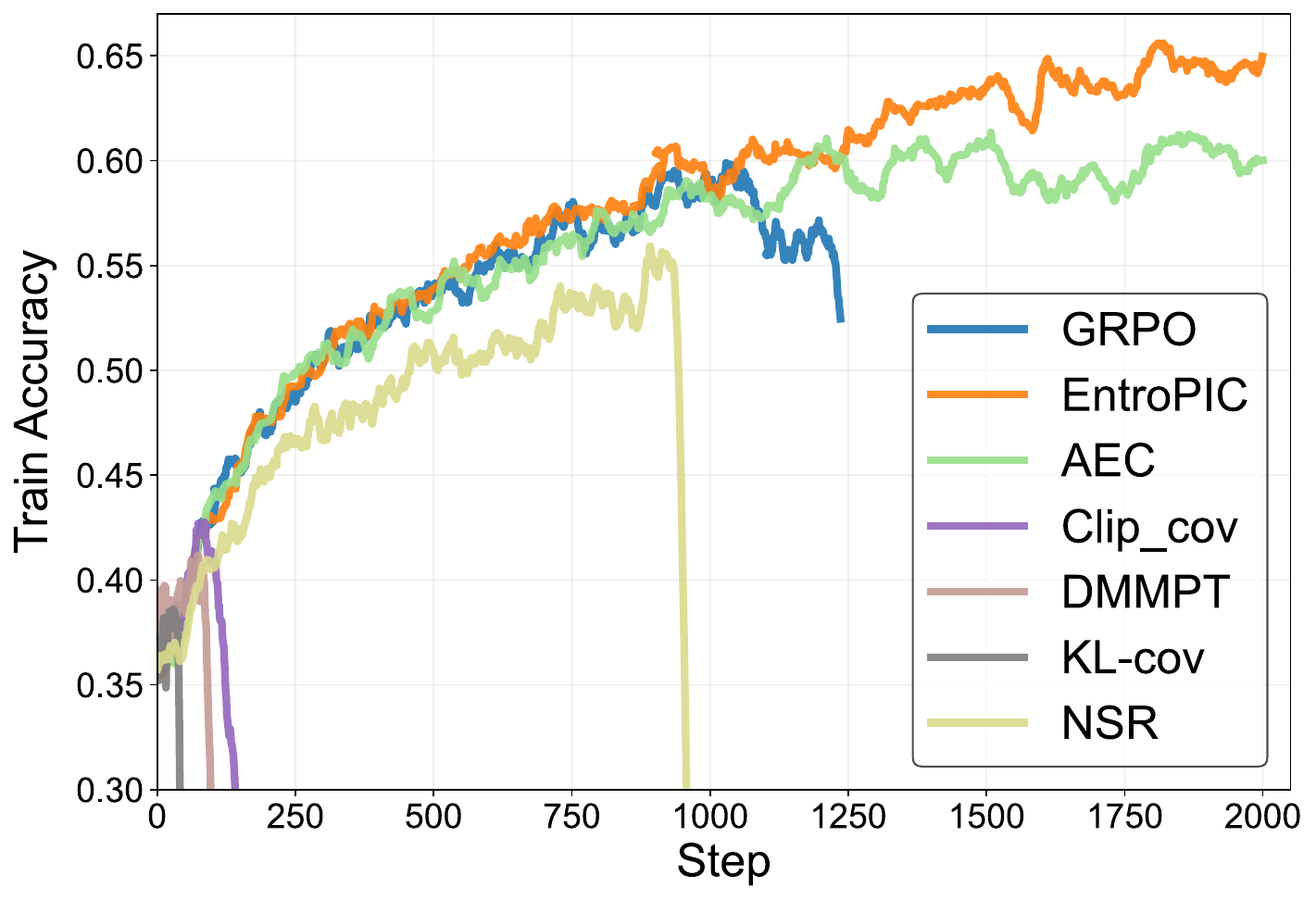}
    \end{minipage}
    }
    \subfigure{
    \begin{minipage}[t]{0.3\linewidth}
    \centering
    \includegraphics[width=\linewidth]{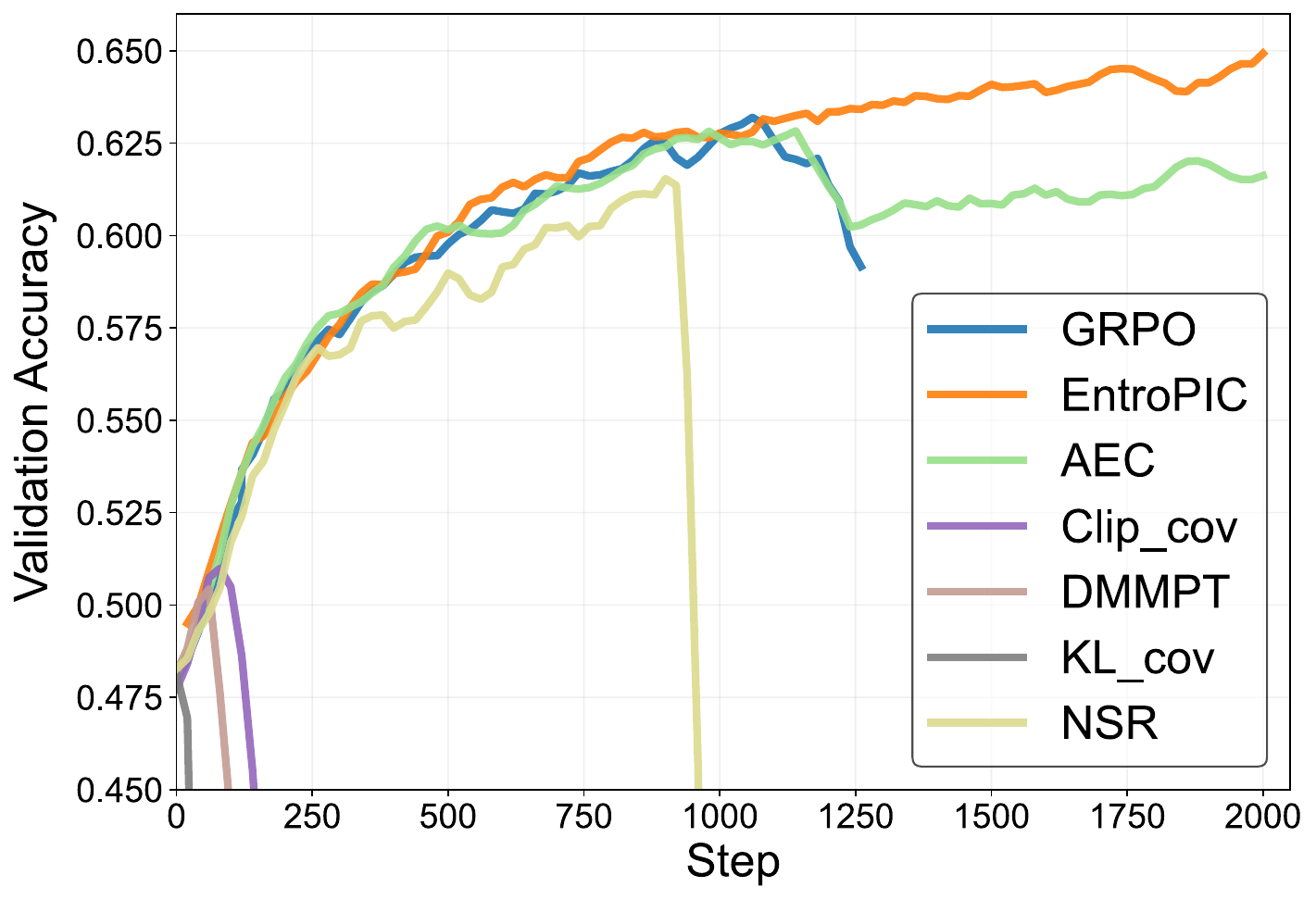}
    \end{minipage}
    }
    % \vspace{-0.5em}
    \caption{\textbf{Left}: Entropy change curves of different methods during training. \textbf{Middle}: Accuracy on the training set. \textbf{Right}: Accuracy on the validation set. EntroPIC stabilizes entropy at the target value throughout training, ensuring steady growth on both training and validation sets, ultimately achieving long-term stable performance.}
    % \vspace{-0.5em}
    \label{exp_on}
\end{figure*}

Following the above verification, we conduct large-scale experiments to assess EntroPIC under various configurations, including on-policy, off-policy, and plug-and-play integration scenarios. We compare our method with GRPO and other baselines in terms of entropy stability, convergence speed across multiple benchmarks. 

\textbf{On-policy Results.} We compare the EntroPIC method with various entropy control approaches in Figure \ref{exp_on}. The baseline demonstrates a significant drop in entropy after 1,000 training steps, leading to a subsequent decline in performance metrics. Methods like Clip\_cov, KL\_cov, and NSR, which are effective in off-policy training, result in entropy increase and failed control during on-policy training, causing instability. While AEC can maintain entropy above the target value, only EntroPIC successfully stabilizes entropy at the target throughout the training process.

In terms of accuracy, other methods show slow improvements after 1,000 steps, while EntroPIC demonstrates consistent performance growth, achieving long-term stability and the best final results. We further evaluate the best-performing checkpoints from methods that behaved normally during on-policy training on the evaluation set. Detailed results are provided in Table \ref{tab_main}. EntroPIC outperforms other methods with a 3.5\% improvement in average pass rate (\texttt{avg@N}) and a 3.8\% improvement in pass rate (\texttt{pass@N}) over GRPO.

\begin{table*}[htbp]
\centering
\caption{Comprehensive Model Performance Comparison (\%). We report results for both standard On-Policy training and Off-Policy settings. The numbers in parentheses indicate \texttt{pass@N} results. EntroPIC consistently demonstrates superior performance across both training paradigms.}
\label{tab_main}
\footnotesize
\renewcommand{\arraystretch}{1.2}
\setlength{\tabcolsep}{4pt}
\resizebox{\textwidth}{!}{
\begin{tabular}{l|cc|cc|cc|cc|cc|cc|cc}
\toprule[1.2pt]
\multirow{2}{*}{\textbf{Models}} & \multicolumn{2}{c|}{\textbf{Math}} & \multicolumn{2}{c|}{\textbf{AMC}} & \multicolumn{2}{c|}{\textbf{AIME24}} & \multicolumn{2}{c|}{\textbf{AIME25}} & \multicolumn{2}{c|}{\textbf{Olympiad}} & \multicolumn{2}{c|}{\textbf{Omni-math}} & \multicolumn{2}{c}{\textbf{Overall}} \\
& \texttt{avg@N} & \texttt{pass@N} & \texttt{avg@N} & \texttt{pass@N} & \texttt{avg@N} & \texttt{pass@N} & \texttt{avg@N} & \texttt{pass@N} & \texttt{avg@N} & \texttt{pass@N} & \texttt{avg@N} & \texttt{pass@N} & \texttt{avg@N} & \texttt{pass@N} \\
\midrule[0.8pt]
\multicolumn{15}{c}{\textit{\textbf{On-Policy Training Settings}}} \\
\midrule[0.5pt]
Initial Model & 86.1 & 97.0 & 58.4 & 81.6 & 23.4 & 60.0 & 23.0 & 53.0 & 49.9 & 68.7 & 32.0 & 49.3 & 45.5 & 68.3 \\
GRPO & 91.2 & 97.4 & 75.1 & 88.0 & 34.3 & 70.0 & 31.0 & 53.3 & 59.1 & 72.7 & 40.7 & 57.6 & 55.2 & 73.2 \\
NSR & 91.5 & 96.4 & 74.1 & 89.2 & 34.7 & 63.3 & 30.0 & 46.7 & 58.5 & 71.3 & 39.7 & 56.2 & 54.8 & 70.5 \\
AEC & \textbf{92.5} & \textbf{97.8} & 77.6 & 89.2 & 37.1 & 73.3 & 31.6 & 60.0 & \textbf{60.9} & \textbf{72.5} & 42.0 & \textbf{58.5} & 56.9 & 75.2 \\
EntroPIC & 92.4 & 97.2 & \textbf{80.1} & \textbf{91.6} & \textbf{42.3} & \textbf{76.7} & \textbf{34.6} & \textbf{66.7} & 60.0 & 71.3 & \textbf{42.7} & 58.4 & \textbf{58.7} & \textbf{77.0} \\
\midrule[0.8pt]
\multicolumn{15}{c}{\textit{\textbf{Off-Policy Training Settings}}} \\
\midrule[0.5pt]
GRPO& 88.7 & 93.6 & 64.0 & 87.9 & 28.9 & 63.3 & 25.5 & 50.0 & 53.2 & 69.0 & 35.3 & 52.4 & 49.3 & 69.4 \\
EntroPIC\_P & 89.8 & 96.4 & 67.8 & \textbf{90.4} & \textbf{34.8} & 66.7 & 27.5 & \textbf{53.3} & 56.4 & 71.1 & 36.6 & 54.9 & 52.2 & 72.2 \\
EntroPIC\_PI & \textbf{91.9} & \textbf{97.0} & \textbf{75.3} & \textbf{90.4} & 34.7 & \textbf{70.0} & \textbf{27.6} & \textbf{53.3} & \textbf{58.8} & \textbf{71.9} & \textbf{40.0} & \textbf{56.8} & \textbf{54.7} & \textbf{73.2} \\
\bottomrule[1.2pt]
\end{tabular}
}
\end{table*}

\begin{figure*}[h]
    \centering
    \subfigure{
    \begin{minipage}[t]{0.3\linewidth}
    \centering
    \includegraphics[width=\linewidth]{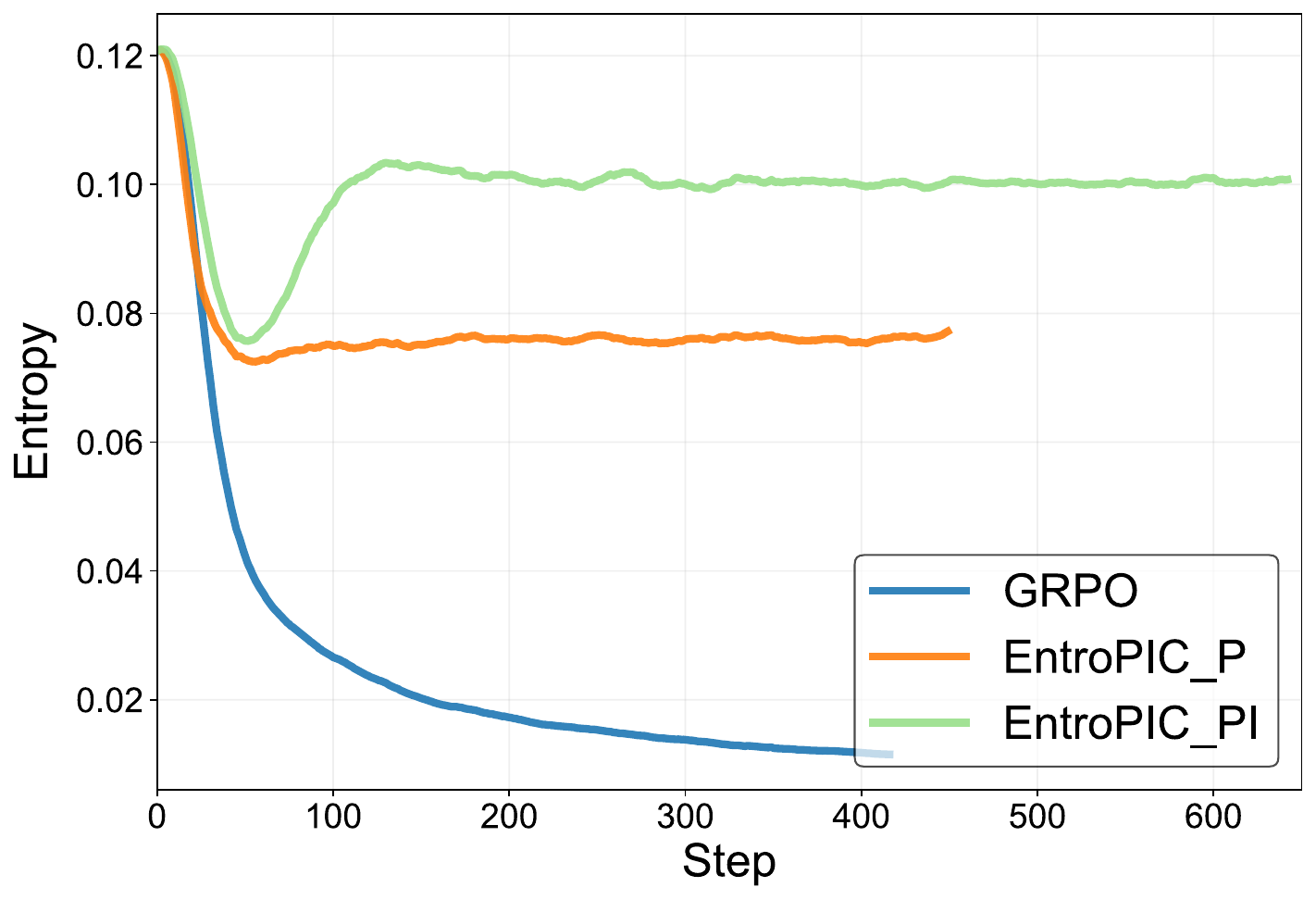}
    \end{minipage}
    }
    \subfigure{
    \begin{minipage}[t]{0.3\linewidth}
    \centering
    \includegraphics[width=\linewidth]{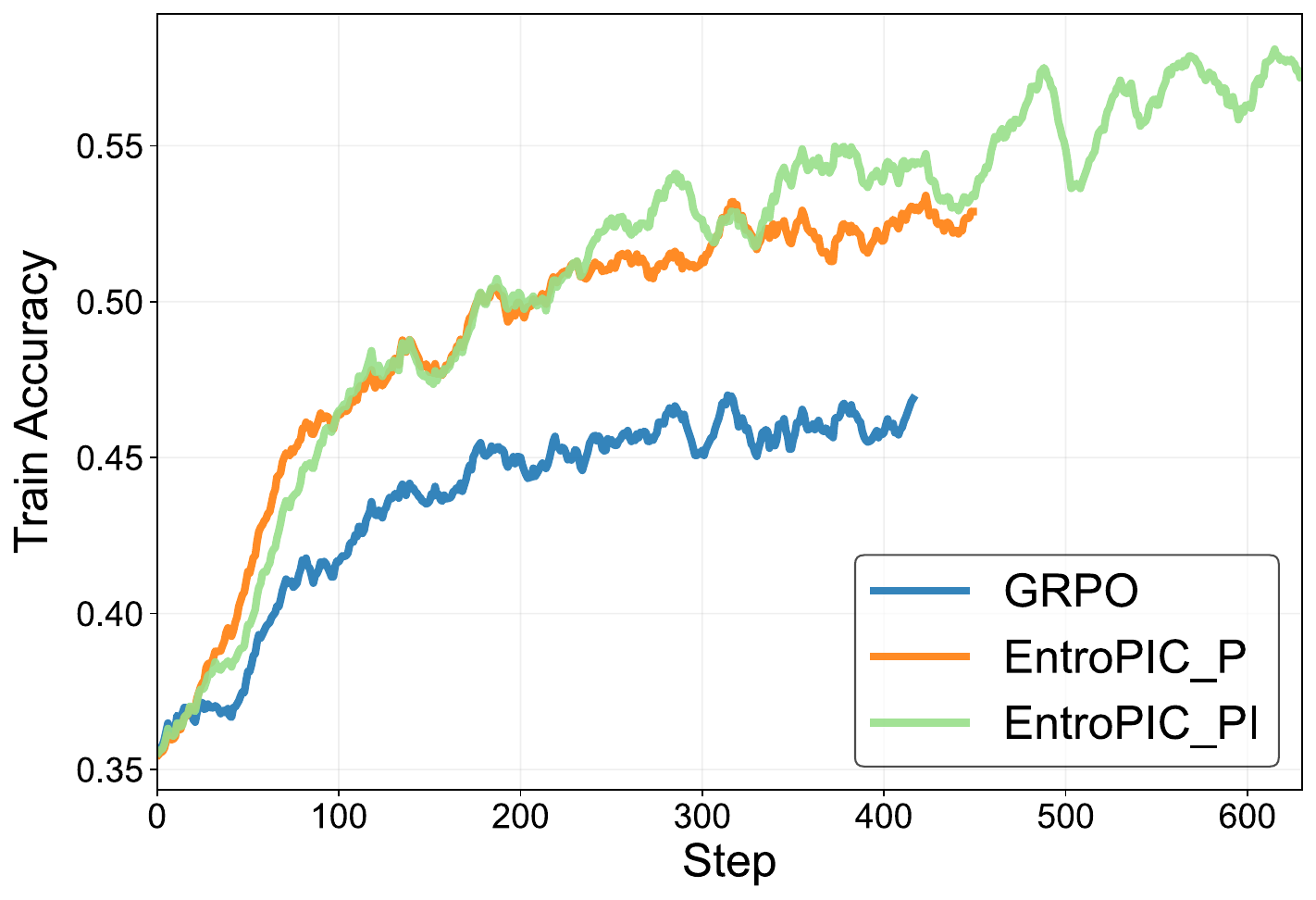}
    \end{minipage}
    }
    \subfigure{
    \begin{minipage}[t]{0.3\linewidth}
    \centering
    \includegraphics[width=\linewidth]{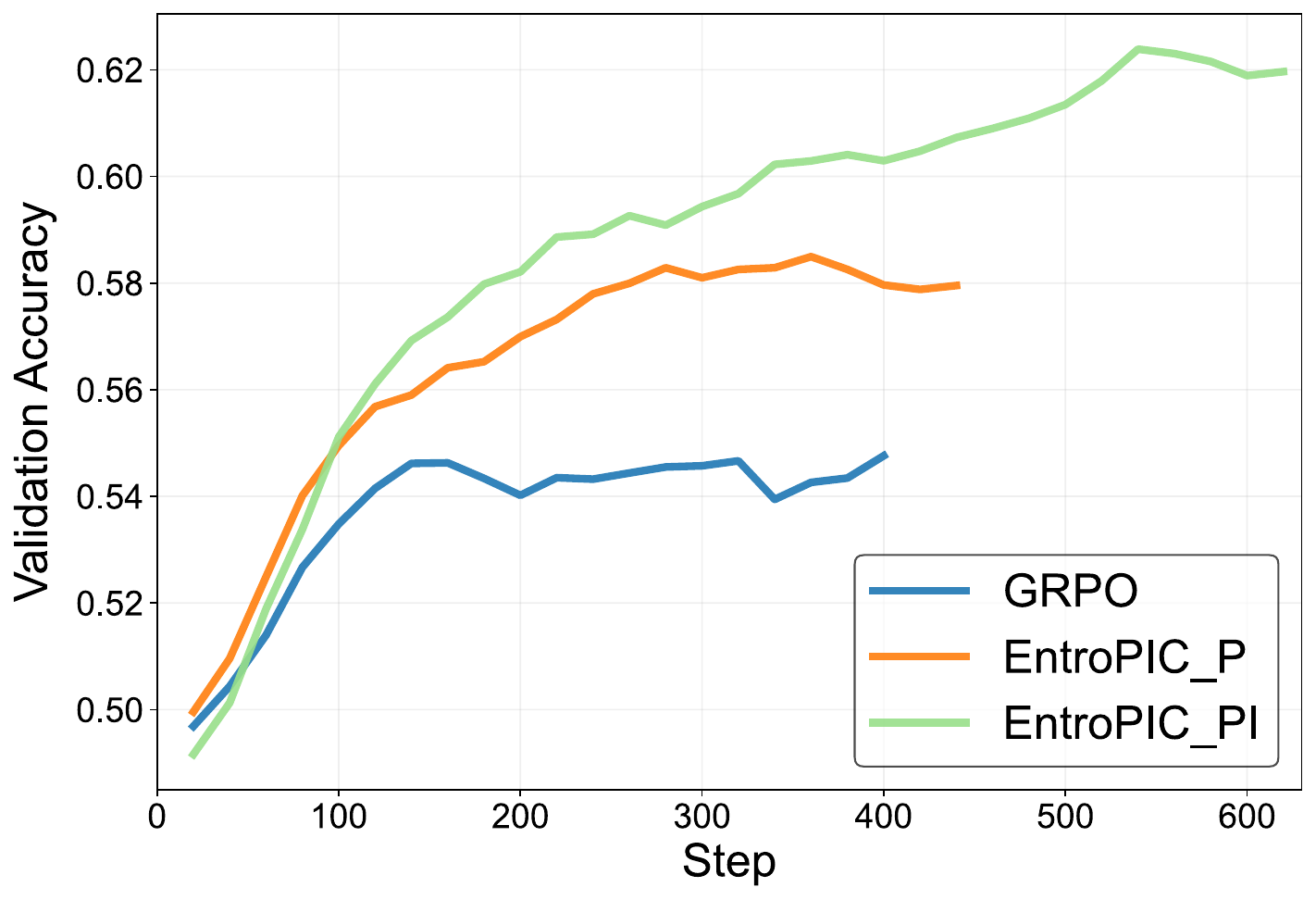}
    \end{minipage}
    }
    % \vspace{-0.5em} 
    \caption{Off-policy training experiment. The entropy of the GRPO method drops significantly, while the EntroPIC method stabilizes entropy at the target value, continuously improving model performance.}
    \label{exp_off}
\end{figure*}

\textbf{Off-policy Results.}
In the off-policy setting, we validate the effectiveness of the EntroPIC method by comparing it with the GRPO approach. Additionally, we conduct a comparative experiment using a P-control-only method. As shown in Figure \ref{exp_off} and Table \ref{tab_main}, EntroPIC with PI control leads to better model performance.

Comparing the on-policy and off-policy metrics, we observe that on-policy training can achieve a higher performance ceiling. Therefore, focusing on entropy-stabilized on-policy training is crucial for unlocking higher final performance.

\textbf{Reasoning Model Results.}
In addition to non-thinking models, we evaluate EntroPIC on reasoning models and compare it against strong baselines trained on the same backbone, including QuestA-Nemotron and JustRL-Nemotron. All models are evaluated on the test set using a diverse collection of mathematical benchmarks.  As shown in Figure \ref{entropic_reasoning} and Table~\ref{tab_think}, EntroPIC consistently outperforms both QuestA-Nemotron and JustRL-Nemotron across most benchmarks, achieving the best overall performance.

\begin{table*}[!htbp]
\centering
\caption{Results on Nemotron-1.5B backbone. All scores except MATH, Minerva, and Olympiad use 32 sampling; those three use 4.}
\label{tab_think}
\footnotesize
\renewcommand{\arraystretch}{1.2}
\setlength{\tabcolsep}{4pt}
\resizebox{\textwidth}{!}{
\begin{tabular}{l|cc|cc|cc|cc|cc|cc|cc|cc|cc|cc}
\toprule[1.2pt]
\textbf{Models} & \multicolumn{2}{c|}{\textbf{Math}} & \multicolumn{2}{c|}{\textbf{AMC}} & \multicolumn{2}{c|}{\textbf{AIME24}} & \multicolumn{2}{c|}{\textbf{AIME25}} & \multicolumn{2}{c|}{\textbf{Olympiad}} & \multicolumn{2}{c|}{\textbf{Minerva}} & \multicolumn{2}{c|}{\textbf{HMMT}} & \multicolumn{2}{c|}{\textbf{BRUMO}} & \multicolumn{2}{c|}{\textbf{CMIMC}}  & \multicolumn{2}{c}{\textbf{Overall}} \\
& \texttt{avg@N} & \texttt{pass@N} & \texttt{avg@N} & \texttt{pass@N} & \texttt{avg@N} & \texttt{pass@N} & \texttt{avg@N} & \texttt{pass@N} & \texttt{avg@N} & \texttt{pass@N} & \texttt{avg@N} & \texttt{pass@N} & \texttt{avg@N} & \texttt{pass@N}& \texttt{avg@N} & \texttt{pass@N} & \texttt{avg@N} & \texttt{pass@N} & \texttt{avg@N} & \texttt{pass@N}  \\
\midrule[0.8pt]
Nemotron-1.5B & 88.7 & 95.4 & 86.6 & \textbf{100} & 51.7 & 83.3 & 46.4 & 73.3 & 62.1 & 75.3 & 25.5 & 36.4 & 30.9 & \textbf{76.7} & 49.6 & 83.3 & 26.8 & 72.5 & 52.0 & 77.4\\
QuestA-Nemotron & 93.2 & 96.8 & 94.1 & \textbf{100} & 72.5 & 83.3 & 63.1 & 83.3 & \textbf{71.1} & \textbf{78.5} & 25.3 & 32.7 & 42.1 & 73.3 & \textbf{70.0} & \textbf{96.7} & 42.1 & 75.0 & 63.7 & 79.9\\
JustRL-Nemotron & \textbf{94.2} & \textbf{97.6} & 95.4 & \textbf{100} & 69.6 & 86.7 & 61.5 & 83.3 & 70.5 & 77.9 & 23.9 & 31.3 & 37.5 & 63.3 & 67.2 & 90.0 & 39.2 & 72.5 & 62.1 & 78.1\\
EntroPIC-Nemotron & 93.2 & 96.8 & \textbf{96.4} & \textbf{100} & \textbf{74.9} & \textbf{90.0} & \textbf{68.3} & \textbf{93.3} & 70.1 & 78.3 & \textbf{36.4} & \textbf{46.7} & \textbf{42.7} & \textbf{76.7} & 63.8 & 93.3 & \textbf{43.0} & \textbf{77.5} & \textbf{65.4} & \textbf{83.6}\\
\bottomrule[1.2pt]
\end{tabular}
}
\end{table*}

%% file: chapters/6_conclusion.tex
\section{Conclusion}
In this paper, we addressed the critical challenge of entropy instability in large language models, a primary bottleneck hindering stable long-term training. We propose EntroPIC, which adjusts the update weights of high-probability samples to stabilize entropy at a predefined target value throughout the training process. The convergence of EntroPIC in entropy control is validated both theoretically and empirically through toy experiments, demonstrating its ability to maintain stability. Experimental results further show that EntroPIC not only achieves more consistent entropy regulation but also facilitates higher performance ceilings compared to existing  state-of-the-art methods. The approach also demonstrates robust generalization across diverse settings, including on-policy training, off-policy training, as well as both non-thinking models and reasoning models.

\textbf{Limitations:} While EntroPIC addresses key issues such as training collapse due to uncontrolled entropy increase and reduced exploration capacity resulting from entropy decline in prolonged training, it is not universally beneficial. In training scenarios where on-policy training alone is sufficient to maintain entropy stability, EntroPIC may not provide significant improvements. Additionally, since the target entropy needs to be manually set, its value must be carefully adjusted to fit the specific requirements of the model and task, which can be a limitation in automated or highly dynamic settings.

%% file: appendix/A_proof.tex
\section{Proof}

\subsection{Proof of Corollary \ref{corollary1}}
\label{proof_co}
According to Theorem \ref{theorem1} in ~\citep{cui2025entropy}, we have 
\begin{equation*}
\mathcal{H}\left(\pi_{\theta}^{k+1} \mid s\right)-\mathcal{H}\left(\pi_{\theta}^{k} \mid s\right) = -\eta \cdot \operatorname{Cov}_{a \sim \pi_{\theta}^{k}(\cdot \mid s)}\left(\log \pi_{\theta}^{k}(a \mid s), \pi_{\theta}^{k}(a \mid s) \cdot A(s, a)\right). 
\end{equation*}
Based on the assumption, for the binary distribution, since all positive and negative samples have the same reward and the expectation of $A(s,a)$ is 0, all positive samples have $A(s,a) > 0$ while all negative samples have $A(s,a) < 0$, and the magnitude of the value is independent of a. Denoting the advantage of positive samples as $A_\text{pos}(s) > 0$ and substituting it into the above formula, we get:
\begin{align*}
\mathcal{H}\left(\pi_{\theta}^{k+1} \mid s\right)-\mathcal{H}\left(\pi_{\theta}^{k} \mid s\right) &= -\eta \cdot \operatorname{Cov}_{a \sim \pi_{\theta}^{k}(\cdot \mid s)}\left(\log \pi_{\theta}^{k}(a \mid s), \pi_{\theta}^{k}(a \mid s) \cdot A_{\text{pos}}(s)\right) \\ &= -\eta \cdot A_{\text{pos}}(s) \cdot \operatorname{Cov}_{a \sim \pi_{\theta}^{k}(\cdot \mid s)}\left(\log \pi_{\theta}^{k}(a \mid s), \pi_{\theta}^{k}(a \mid s) \right) \\&\le 0.
\end{align*}
Therefore, positive samples will definitely lead to a decrease in entropy; similarly, negative samples will definitely lead to an increase in entropy.

\subsection{Proof of Theorem \ref{theorem1}}
\label{proof_theorem1}

First, we define a coefficient function based on the advantage value:
\begin{align*} 
c(A(s,a')) &= 
\begin{cases}  
1+\alpha & \text{if } A(s,a') > 0, \\ 
1-\alpha & \text{if } A(s,a') < 0. 
\end{cases} 
\end{align*}

With this definition, we derive the gradient of the loss function $\nabla_{\theta_{s,a}} L(\theta)$ and the parameter update $\Delta\theta_{s,a}^k$ step by step:
\begin{align*} 
\nabla_{\theta_{s,a}} L(\theta) 
&= \nabla_{\theta_{s,a}} \left( - \mathbb{E}_{a' \sim \pi_\theta(\cdot|s)} \left[ c(A(s,a')) \cdot A(s,a') \cdot \log \pi_\theta(a'|s) \right] \right) \\ 
&= - \mathbb{E}_{a' \sim \pi_\theta(\cdot|s)} \left[ c(A(s,a')) \cdot A(s,a') \cdot \nabla_{\theta_{s,a}} \log \pi_\theta(a'|s) \right] \\  
&= - \sum_{a' \in \mathcal{A}} \pi_\theta(a'|s) \cdot c(A(s,a')) \cdot A(s,a') \cdot \frac{\partial \log \pi_\theta(a'|s)}{\partial \theta_{s,a}} \\  
&= - \sum_{a' \in \mathcal{A}} \pi_\theta(a'|s) \cdot c(A(s,a')) \cdot A(s,a') \cdot \left( \mathbb{I}(a'=a) - \pi_\theta(a|s) \right) \\  
&= - \left[ \pi_\theta(a|s) \cdot c(A(s,a)) \cdot A(s,a) \cdot (1 - \pi_\theta(a|s)) \right. \\ 
&\quad \left. + \sum_{a' \neq a} \pi_\theta(a'|s) \cdot c(A(s,a')) \cdot A(s,a') \cdot (-\pi_\theta(a|s)) \right] \\ 
&= \pi_\theta(a|s) \cdot \left[ \sum_{a' \in \mathcal{A}} \pi_\theta(a'|s) \cdot c(A(s,a')) \cdot A(s,a') - c(A(s,a)) \cdot A(s,a) \right] \\  
\Delta\theta_{s,a}^k 
&=- \eta \cdot \nabla_{\theta_{s,a}} L(\theta) \\ 
&= \eta \cdot \pi_\theta(a|s) \cdot \left( -\sum_{a' \in \mathcal{A}} \pi_\theta(a'|s) \cdot c(A(s,a')) \cdot A(s,a') + c(A(s,a)) \cdot A(s,a) \right). 
\end{align*}

Next, we calculate the the entropy change direction $\left\langle\nabla_{\theta} \mathcal{H}\left(\theta^{k} \mid s\right), \Delta \theta^{k}\right\rangle$ and simplify it:
\begin{align*}
&\left\langle\nabla_{\theta} \mathcal{H}\left(\theta^{k} \mid s\right), \Delta \theta^{k}\right\rangle 
= -\mathbb{E}_{a \sim \pi_{\theta}(\cdot \mid s)}\left[ \log \pi_{\theta}(a \mid s) \cdot \eta \cdot \pi_{\theta}(a \mid s) \right. \\
&\quad \left. \cdot \left( -\sum_{a' \in \mathcal{A}} \pi_{\theta}(a' \mid s) \cdot c(A(s,a')) \cdot A(s,a') + c(A(s,a)) \cdot A(s,a) \right) \right] \\
&\quad + \mathbb{E}_{a \sim \pi_{\theta}(\cdot \mid s)}\left[\log \pi_{\theta}(a \mid s)\right] \cdot \mathbb{E}_{a^{\prime} \sim \pi_{\theta}(\cdot \mid s)}\left[ \eta \cdot \pi_{\theta}(a' \mid s) \right. \\
&\quad \left. \cdot \left( -\sum_{a'' \in \mathcal{A}} \pi_{\theta}(a'' \mid s) \cdot c(A(s,a'')) \cdot A(s,a'') + c(A(s,a')) \cdot A(s,a') \right) \right] \\
&= \eta \sum_{a \in \mathcal{A}} \pi_{\theta}(a \mid s) \cdot \log \pi_{\theta}(a \mid s) \cdot \pi_{\theta}(a \mid s) \cdot \left( \sum_{a' \in \mathcal{A}} \pi_{\theta}(a' \mid s) \cdot c(A(s,a')) \cdot A(s,a') - c(A(s,a)) \cdot A(s,a) \right) \\
&\quad + \eta \cdot \left( \sum_{a \in \mathcal{A}} \pi_{\theta}(a \mid s) \cdot \log \pi_{\theta}(a \mid s) \right) \cdot \sum_{a' \in \mathcal{A}} \pi_{\theta}(a' \mid s) \cdot \pi_{\theta}(a' \mid s) \\
&\quad \cdot \left( -\sum_{a'' \in \mathcal{A}} \pi_{\theta}(a'' \mid s) \cdot c(A(s,a'')) \cdot A(s,a'') + c(A(s,a')) \cdot A(s,a') \right) \\
&= \eta \sum_{a \in \mathcal{A}} \pi_{\theta}^2(a \mid s) \cdot \log \pi_{\theta}(a \mid s) \cdot \sum_{a' \in \mathcal{A}} \pi_{\theta}(a' \mid s) \cdot c(A(s,a')) \cdot A(s,a') \\
&\quad - \eta \sum_{a \in \mathcal{A}} \pi_{\theta}^2(a \mid s) \cdot \log \pi_{\theta}(a \mid s) \cdot c(A(s,a)) \cdot A(s,a) \\
&\quad - \eta \cdot \left( \sum_{a \in \mathcal{A}} \pi_{\theta}(a \mid s) \cdot \log \pi_{\theta}(a \mid s) \right)  \cdot \sum_{a'' \in \mathcal{A}} \pi_{\theta}(a'' \mid s) \cdot c(A(s,a'')) \cdot A(s,a'')  \cdot \sum_{a' \in \mathcal{A}} \pi_{\theta}^2(a' \mid s) \\
&\quad + \eta \cdot \left( \sum_{a \in \mathcal{A}} \pi_{\theta}(a \mid s) \cdot \log \pi_{\theta}(a \mid s) \right) \cdot \sum_{a' \in \mathcal{A}} \pi_{\theta}^2(a' \mid s) \cdot c(A(s,a')) \cdot A(s,a') \\
&= \eta \sum_{a' \in \mathcal{A}} \pi_{\theta}(a' \mid s) \cdot c(A(s,a')) \cdot A(s,a') \cdot \sum_{a \in \mathcal{A}} \pi_{\theta}^2(a \mid s) \cdot \log \pi_{\theta}(a \mid s) \\
&\quad - \eta \sum_{a \in \mathcal{A}} \pi_{\theta}^2(a \mid s) \cdot \log \pi_{\theta}(a \mid s) \cdot c(A(s,a)) \cdot A(s,a) \\
&\quad - \eta \sum_{a'' \in \mathcal{A}} \pi_{\theta}(a'' \mid s) \cdot c(A(s,a'')) \cdot A(s,a'') \cdot \sum_{a' \in \mathcal{A}} \pi_{\theta}^2(a' \mid s) \\
&\quad + \eta \sum_{a' \in \mathcal{A}} \pi_{\theta}^2(a' \mid s) \cdot c(A(s,a')) \cdot A(s,a') \cdot \sum_{a \in \mathcal{A}} \pi_{\theta}(a \mid s) \cdot \log \pi_{\theta}(a \mid s),
\end{align*}
where we use the identity: 
$$
\log \pi_{\theta}(a|s) - \sum_{a'} \pi_{\theta}(a' | s) \log \pi_{\theta}(a'|s) = \sum_{a'} \pi_{\theta}(a'|s) \left( \log \pi_{\theta}(a \mid s) - \log \pi_{\theta}(a' \mid s) \right).
$$

First, rewrite Term (III) by renaming dummy variables (consistent with Term (I)):
$$
- \eta \sum_{a' \in \mathcal{A}} \pi_{\theta}(a' \mid s) \cdot c(A(s,a')) \cdot A(s,a') \cdot \sum_{a \in \mathcal{A}} \pi_{\theta}^2(a \mid s) \cdot \sum_{a'' \in \mathcal{A}} \pi_{\theta}(a'' \mid s) \cdot \log \pi_{\theta}(a'' \mid s).  \quad \text{(V)}
$$

Combine Term (I) and Term (V):
\begin{align*}
&\eta\sum_{a'\in\mathcal{A}}\pi_\theta(a'|s)c(A(s,a'))A(s,a')\sum_{a\in\mathcal{A}}\pi_\theta^2(a|s)\sum_{a''\in\mathcal{A}}\pi_\theta(a''|s)\left[\log\pi_\theta(a|s)-\log\pi_\theta(a''|s)\right] \\
&= \eta\sum_{a\in\mathcal{A}}\pi_\theta(a|s) c(A(s,a)) A(s,a)\sum_{a'\in\mathcal{A}}\pi_\theta^2(a'|s)\sum_{a''\in\mathcal{A}}\pi_\theta(a''|s)\left[\log\pi_\theta(a'|s) - \log\pi_\theta(a''|s)\right]. 
\end{align*}

Similarly, combine Term (II) and Term (IV):
\begin{align*}
&-\eta\sum_{a'\in\mathcal{A}}\pi^2_\theta(a'|s)c(A(s,a'))A(s,a')\sum_{a\in\mathcal{A}}\pi_\theta(a|s)\left[\log\pi_\theta(a'|s)-\log\pi_\theta(a|s)\right] \\ 
&= -\eta\sum_{a\in\mathcal{A}}\pi_\theta^2(a|s) c(A(s,a)) A(s,a)\sum_{a'\in\mathcal{A}}\pi_\theta(a'|s)\left[\log\pi_\theta(a|s) - \log\pi_\theta(a'|s)\right].
\end{align*}

Adding the two combined results and simplifying further, we obtain:
\begin{align*}
\left\langle\nabla_{\theta} \mathcal{H}\left(\theta^{k} \mid s\right), \Delta \theta^{k}\right\rangle 
&=\eta \sum_{a} \pi_{\theta}(a \mid s) c(A(s,a)) A(s,a) \sum_{a'} \pi_{\theta}^2(a' \mid s) \cdot\\ & \quad\quad \left[ \log \pi_{\theta}(a' \mid s) - \sum_{a''} \pi_{\theta}(a'' \mid s) \log \pi_{\theta}(a'' \mid s) \right] \\
&\quad - \eta \sum_{a} \pi_{\theta}^2(a \mid s) c(A(s,a)) A(s,a) \left[ \log \pi_{\theta}(a \mid s) - \sum_{a'} \pi_{\theta}(a' \mid s) \log \pi_{\theta}(a' \mid s) \right] \\ 
&= -\eta \sum_{a\in\mathcal{A}} \sum_{a^{\prime}\in\mathcal{A}} c(A(s, a)) \cdot A(s, a) \cdot \pi_{\theta}(a \mid s) \cdot \pi_{\theta}\left(a^{\prime} \mid s\right) \\
&\quad \cdot\left(\pi_{\theta}(a \mid s)-\pi_{\theta}\left(a^{\prime} \mid s\right)\right) \left(\log \pi_{\theta}(a \mid s)-\log \pi_{\theta}\left(a^{\prime} \mid s\right)\right)\\
&= -\eta \cdot (1+\alpha) \sum_{a,a' \in \mathcal{A}_{\text{pos}}} A_{\text{pos}}(s) \cdot \pi_{\theta}^2(a \mid s) \pi_{\theta}(a' \mid s) \left( \log \pi_{\theta}(a \mid s) - \log \pi_{\theta}(a' \mid s) \right) \\
&\quad -\eta \cdot (1-\alpha) \sum_{a,a' \in \mathcal{A}_{\text{neg}}} A_{\text{neg}}(s) \cdot \pi_{\theta}^2(a \mid s) \pi_{\theta}(a' \mid s) \left( \log \pi_{\theta}(a \mid s) - \log \pi_{\theta}(a' \mid s) \right)\\
&= -\eta(1+\alpha)A_{\text{pos}}(s) S_{\text{pos}} - \eta(1-\alpha)A_{\text{neg}}(s) S_{\text{neg}},
\end{align*}
where $\mathcal{A}_{\text{pos}}(s)$ (resp. $\mathcal{A}_{\text{neg}}(s)$) denotes the set of actions with positive (resp. negative) advantage, $A_{\text{pos}}(s) > 0$ (resp. $A_{\text{neg}}(s) < 0$) is the constant advantage value for positive (resp. negative) actions, and:
$S_{\text{pos}} = \sum_{a,a' \in \mathcal{A}_{\text{pos}}} \pi_{\theta}^2(a \mid s) \pi_{\theta}(a' \mid s) \left( \log \pi_{\theta}(a \mid s) - \log \pi_{\theta}(a' \mid s) \right) \geq 0$,
$S_{\text{neg}} = \sum_{a,a' \in \mathcal{A}_{\text{neg}}} \pi_{\theta}^2(a \mid s) \pi_{\theta}(a' \mid s) \left( \log \pi_{\theta}(a \mid s) - \log \pi_{\theta}(a' \mid s) \right) \geq 0$.

Since the sum of advantages equals zero, we let $A_{\text{neg}} = -hA_{\text{pos}}$ (where $h > 0$ is a positive constant). Substituting this into the above equation gives:
$$
\left\langle\nabla_{\theta} \mathcal{H}\left(\theta^{k} \mid s\right), \Delta \theta^{k}\right\rangle  = -\eta A_{\text{pos}}(s) \left[ (1 + \alpha) S_{\text{pos}} - h (1 - \alpha) S_{\text{neg}} \right].
$$

When the entropy stabilizes, the error $e_k = 0$ and $\alpha = 0$, so:
$$
\left\langle\nabla_{\theta} \mathcal{H}\left(\theta^{k} \mid s\right), \Delta \theta^{k}\right\rangle=\eta A_{\text{pos}}(s)(S_{\text{pos}} - hS_{\text{neg}})=0.
$$

\textbf{P Control.} Let the entropy error be $e_k = \mathcal{H}(\theta^k|s)-\mathcal{H}_{\text{tar}}$ (where $\mathcal{H}_{\text{tar}}$ is the target entropy). For proportional control (P control), we set $\alpha = K_p e_k$ (where $K_p > 0$ is the proportional gain). Substituting $\alpha$ into the inner product formula gives:
$$
\left\langle\nabla_{\theta} \mathcal{H}\left(\theta^{k} \mid s\right), \Delta \theta^{k}\right\rangle  = C_0 + C_1 \cdot e_k,
$$
where:
 $C_0 = \eta A_{\text{pos}}(S_{\text{pos}} - hS_{\text{neg}}) = 0$ (from entropy stabilization condition),
 $C_1 = -\eta A_{\text{pos}} K_p (S_{\text{pos}} + hS_{\text{neg}}) < 0$ (all terms are positive except the negative sign).

The error at the next step satisfies:
$$
|e_{k+1}| = \left| e_k + \Delta\mathcal{H}(\theta_k|s) \right| = \left| e_k + C_1 e_k \right| = \left| (1 + C_1) e_k \right| < e_k.
$$
since $0 < 1 + C_1 < 1$. Thus, the error converges to zero:
$$
\lim_{k \to \infty} |e_k| = \lim_{k \to \infty} |1 + C_1|^k |e_0| \to 0.
$$

\textbf{PI Control.} For PI control, we set $\alpha = K_p e_k + K_i I_k$, where:
 $e_k = \mathcal{H}(\theta^k|s) - \mathcal{H}_{\text{tar}}$,
 $I_k = \sum_{t=1}^{k-1} e_t$,
 $K_p > 0$ (proportional gain) and $K_i > 0$.

From the previous derivation, the entropy change can be written as:
$$
\left\langle\nabla_{\theta} \mathcal{H}\left(\theta^{k} \mid s\right), \Delta \theta^{k}\right\rangle = -C (K_p e_k + K_i I_k),
$$
where $C = \eta A_{\text{pos}} (S_{\text{pos}} + hS_{\text{neg}}) > 0$.

The error evolution satisfies $\Delta\mathcal{H}(\theta_k|s) = \left\langle \nabla_{\theta} \mathcal{H}\left(\theta^{k} \mid s\right), \Delta \theta^{k}\right\rangle$, so we have:
$$e_{k+1} = e_k + \Delta\mathcal{H}(\theta_k|s) = e_k - C(K_p e_k + K_i I_k).$$

With the integral definition $I_{k+1} = I_k + e_{k+1}$, we get the two-dimensional linear recurrence system:
$$
\begin{bmatrix} e_{k+1} \\ I_{k+1} \end{bmatrix} = \begin{bmatrix} 1 - C K_p & -C K_i \\ 1 - C K_p & 1 - C K_i \end{bmatrix} \begin{bmatrix} e_k \\ I_k \end{bmatrix} = \begin{bmatrix} a & -b \\ a & 1 - b \end{bmatrix} \begin{bmatrix} e_k \\ I_k \end{bmatrix},
$$
where we define simplified coefficients:
$a = 1 - C K_p$, $b = C K_i$.

To analyze the stability of this recurrence system, we construct a Lyapunov function:
$$
V_k = e_k^2 + \frac{b}{1 - b} I_k^2,
$$
where $0 < b < 1$ (to be proven later) ensures the coefficient is positive. Calculating $V_{k+1}$ using the recurrence relation:
$$
V_{k+1} = (a e_k - b I_k)^2 + \frac{b}{1 - b} \left[ a e_k + (1 - b) I_k \right]^2.
$$

We now compute the difference $V_{k+1} - V_k$ correctly:
\begin{align*}
V_{k+1} - V_k 
&= (a e_k - b I_k)^2 + \frac{b}{1 - b} \left[ a e_k + (1 - b) I_k \right]^2 - \left( e_k^2 + \frac{b}{1 - b} I_k^2 \right) \\
% Expand squared terms with correct algebra
&= a^2 e_k^2 - 2ab e_k I_k + b^2 I_k^2 + \frac{b}{1-b}\left(a^2 e_k^2 + 2a(1-b)e_k I_k + (1-b)^2 I_k^2\right) - e_k^2 - \frac{b}{1-b}I_k^2 \\
% Distribute terms properly
&= a^2 e_k^2 - 2ab e_k I_k + b^2 I_k^2 + \frac{a^2 b}{1-b}e_k^2 + 2ab e_k I_k + b(1-b)I_k^2 - e_k^2 - \frac{b}{1-b}I_k^2 \\
% Cancel cross terms and group like terms
&= e_k^2 \left(a^2 + \frac{a^2 b}{1-b} - 1\right) + I_k^2 \left(b^2 + b(1-b) - \frac{b}{1-b}\right) \\
% Correct simplification of coefficients
&= e_k^2 \left( \frac{a^2(1-b) + a^2 b - (1-b)}{1-b} \right) + I_k^2 \left( \frac{b^2(1-b) + b(1-b)^2 - b}{1-b} \right) \\
&= \frac{1}{1-b} \left[ e_k^2 \left(a^2 - (1-b)\right) + I_k^2 \left(b^2 - b^3 + b - 2b^2 + b^3 - b\right) \right] \\
&= \frac{1}{1-b} \left[ e_k^2 \left(a^2 - (1-b)\right) - b^2 I_k^2 \right].
\end{align*}

For stability, we need $V_{k+1} < V_k$, which requires both terms in the brackets to be negative:
1. $a^2 - (1-b) < 0 \implies a^2 < 1-b$
   Substituting back $a = 1 - CK_p$ and $b = CK_i$:
   $$(1 - CK_p)^2 < 1 - CK_i \implies C^2 K_p^2 - 2CK_p + CK_i < 0.$$

2. $-b^2 < 0$, which is always true since $b > 0$
   Additionally, the denominator requires $1 - b > 0 \implies b < 1 \implies CK_i < 1.$

These conditions are easily satisfied with small learning rate $\eta$ (making $C$ small) and moderate gains $K_p, K_i$. Under these conditions:
$$
e_{k+1}^2 + \frac{b}{1-b}I_{k+1}^2 < e_k^2 + \frac{b}{1-b}I_k^2.
$$

Rearranging using $I_{k+1} = I_k + e_{k+1}$:
$$
e_{k+1}^2 - e_k^2 < \frac{b}{1-b}\left(I_k^2 - I_{k+1}^2\right) = -\frac{b}{1-b}e_{k+1}(2I_k + e_{k+1}).
$$

\textbf{Proof by Contradiction (Convergence of $e_k$)}

\textbf{Assumption}: There exists an infinite subsequence $\{k_j\}$ with $|e_{k_j+1}| \geq |e_{k_j}|$.

Since $V_k$ is monotonically decreasing and bounded below by 0, it converges to $V_\infty \geq 0$. Thus:
- $e_k$ is bounded: $|e_k| \leq \sqrt{V_0}$
- $I_k$ is bounded: $|I_k| \leq \sqrt{\frac{1-b}{b}V_0}$

\textbf{Case 1:} $e_{k+1}$ and $I_k$ have the same sign. Here $e_{k+1}I_k > 0$, so $2I_k + e_{k+1}$ has the same sign as $e_{k+1}$. This gives:
$$e_{k+1}(2I_k + e_{k+1}) > 0 \implies -\frac{b}{1-b}e_{k+1}(2I_k + e_{k+1}) < 0.$$
Substituting into our inequality: $e_{k+1}^2 - e_k^2 < 0 \implies |e_{k+1}| < |e_k|$, contradicting the assumption.

\textbf{Case 2:} $e_{k+1}$ and $I_k$ have opposite signs. For the assumption $|e_{k+1}| \geq |e_k|$ to hold, we need:
$$e_{k+1}(2I_k + e_{k+1}) < 0 \quad (*).$$

From $(*)$: (1) If $e_{k+1} > 0$, then $2I_k + e_{k+1} < 0 \implies I_k < -e_{k+1}/2.$ (2) If $e_{k+1} < 0$, then $2I_k + e_{k+1} > 0 \implies I_k > -e_{k+1}/2.$

In both subcases, $|I_k| > |e_{k+1}|/2$, so:
$$|I_{k+1}| = |I_k + e_{k+1}| \leq |I_k| - |e_{k+1}|/2 < |I_k|.$$
Thus $|I_k|$ is monotonically decreasing along $\{k_j\}$.

Since $V_k \to V_\infty$ and $|I_k|$ is decreasing/bounded, $I_k \to I_\infty$. For the recurrence system:
$$\begin{bmatrix} e_\infty \\ I_\infty \end{bmatrix} = \begin{bmatrix} a & -b \\ a & 1-b \end{bmatrix}\begin{bmatrix} e_\infty \\ I_\infty \end{bmatrix}.$$

The only solution is $e_\infty = 0, I_\infty = 0$ under our stability conditions, contradicting the assumption. Thus $\lim_{k \to \infty} e_k = 0$, meaning the entropy converges to the target value.

\subsection{Proof of Theorem \ref{theorem2}}
\label{proof_theorem2}
In the off-policy setting with a PPO-like loss (without clipping), the objective function is defined as:
\begin{align*}
L(\theta) = -\mathbb{E}_{a \sim \mu} \left[ \min \left(\rho(a|s) \cdot A(s,a), \text{clip}(\rho(a|s),1-\epsilon_\text{low},1 + \epsilon_\text{high})\right)A(s,a) \right],
\end{align*}
where $\rho(a|s) = \frac{\pi_\theta(a|s)}{\mu(a|s)}$ is the importance sampling ratio. For simplicity, we define 
\begin{align*}
     \rho_{\text{clip}}(a|s) = 
  \begin{cases} 
  1-\epsilon_{\text{low}} & \text{if } \rho(a|s) < 1-\epsilon_{\text{low}}, \\
  \rho(a|s) & \text{if } 1-\epsilon_{\text{low}} \leq \rho(a|s) \leq 1+\epsilon_{\text{high}}, \\
  1+\epsilon_{\text{high}} & \text{if } \rho(a|s) > 1+\epsilon_{\text{high}.}
  \end{cases}
\end{align*}
The gradient of this loss with respect to parameters $\theta_{s,a}$ is:
\begin{align*}
\nabla_{\theta_{s,a}}& L(\theta) 
= \nabla_{\theta_{s,a}} \left( - \mathbb{E}_{a' \sim \mu(\cdot|s)} \left[ c(A(s,a')) \cdot A(s,a') \cdot \rho_\text{clip}(a'|s) \right] \right) \\
&= \nabla_{\theta_{s,a}} \left( - \mathbb{E}_{a' \sim \mu(\cdot|s)} \left[ c(A(s,a')) \cdot A(s,a') \cdot (1-\epsilon_\text{low}))\mathbb{I}(\rho(a'|s)< 1-\epsilon_\text{low}) \right] \right)\\&+ \nabla_{\theta_{s,a}} \left( - \mathbb{E}_{a' \sim \mu(\cdot|s)} \left[ c(A(s,a')) \cdot A(s,a') \cdot \rho(a'|s)\mathbb{I}(1-\epsilon_\text{low} < \rho(a'|s)< 1+\epsilon_\text{high}) \right] \right)\\ & + \nabla_{\theta_{s,a}} \left( - \mathbb{E}_{a' \sim \mu(\cdot|s)} \left[ c(A(s,a')) \cdot A(s,a') \cdot (1+\epsilon_\text{low}))\mathbb{I}(\rho(a'|s)> 1+\epsilon_\text{low}) \right] \right)\\
&= - \mathbb{E}_{a' \sim \mu(\cdot|s)} \left[ \mathbb{I}(1-\epsilon_\text{low} < \rho(a'|s)< 1+\epsilon_\text{high})\cdot\rho(a'|s) \cdot c(A(s,a')) \cdot A(s,a') \cdot \nabla_{\theta_{s,a}} \log \pi_\theta(a'|s) \right] \\
&= - \sum_{a' \in \mathcal{A}} \mathbb{I}(1-\epsilon_\text{low} < \rho(a'|s)< 1+\epsilon_\text{high})\cdot\pi_\theta(a'|s) \cdot c(A(s,a')) \cdot A(s,a') \cdot \frac{\partial \log \pi_\theta(a'|s)}{\partial \theta_{s,a}} \\
&= - \sum_{a' \in \mathcal{A}} \mathbb{I}(1-\epsilon_\text{low} < \rho(a'|s)< 1+\epsilon_\text{high}) \cdot \pi_\theta(a'|s) \cdot c(A(s,a')) \cdot A(s,a') \cdot \left( \mathbb{I}(a'=a) - \pi_\theta(a|s) \right) \\
&= - \left[ \pi_\theta(a|s) \cdot c(A(s,a)) \cdot A(s,a) \cdot (1 - \pi_\theta(a|s)) \right. \\
&\quad \left. + \sum_{a' \neq a} \mathbb{I}(1-\epsilon_\text{low} < \rho(a'|s)< 1+\epsilon_\text{high})\cdot \pi_\theta(a'|s) \cdot c(A(s,a')) \cdot A(s,a') \cdot (-\pi_\theta(a|s)) \right] \\
&= \pi_\theta(a|s) \left[ \sum_{a' \in \mathcal{A}} \mathbb{I}(1-\epsilon_\text{low} < \rho(a'|s)< 1+\epsilon_\text{high}) \cdot \pi_\theta(a'|s) \cdot c(A(s,a')) \cdot A(s,a') - c(A(s,a)) \cdot A(s,a) \right].
\end{align*}

The parameter update is then:
\begin{align*}
\Delta\theta_{s,a}^k 
&= -\eta \nabla_{\theta_{s,a}} L(\theta) \\
&= \eta \pi_\theta(a|s)\left[ c(A(s,a)) \cdot A(s,a) - \sum_{a' \in \mathcal{A}} \mathbb{I}(1-\epsilon_\text{low} < \rho(a'|s)< 1+\epsilon_\text{high})\cdot\pi_\theta(a'|s) \cdot c(A(s,a')) \cdot A(s,a') \right].
\end{align*}

The directional change of entropy along the update direction is given by the inner product:
\begin{align*}
\left\langle\nabla_{\theta} \mathcal{H}\left(\theta^{k} \mid s\right), \Delta \theta^{k}\right\rangle 
&= -\mathbb{E}_{a \sim \pi_{\theta}(\cdot \mid s)}\left[ \log \pi_{\theta}(a \mid s) \cdot \Delta\theta_{s,a}^k \right] \\
&\quad + \mathbb{E}_{a \sim \pi_{\theta}(\cdot \mid s)}\left[ \log \pi_{\theta}(a \mid s) \right] \cdot \mathbb{E}_{a' \sim \pi_{\theta}(\cdot \mid s)}\left[ \Delta\theta_{s,a'}^k \right].
\end{align*}

Substituting the expression for $\Delta\theta_{s,a}^k$ and simplifying (following similar algebraic manipulations as in the on-policy case but with off-policy corrections), we obtain:
\begin{align*}
\left\langle\nabla_{\theta} \mathcal{H}\left(\theta^{k} \mid s\right), \Delta \theta^{k}\right\rangle 
&= -\eta(1+\alpha) A_{\text{pos}} S_{\text{pos}} - \eta(1-\alpha) A_{\text{neg}} S_{\text{neg}} + \eta \delta \cdot C,
\end{align*}
where:
\begin{itemize}
\item $\mathcal{A}_{\text{pos}}$ and $\mathcal{A}_{\text{neg}}$ denote sets of actions with positive and negative advantages, respectively,
\item $A_{\text{pos}}(s) > 0$ and $A_{\text{neg}}(s) < 0$ are constant advantage values for positive and negative actions,
\item $S_{\text{pos}} = \sum_{a \in \mathcal{A}_{\text{pos}}} \sum_{a' \in \mathcal{A}} \pi_{\theta}^2(a|s) \mu(a'|s) \rho(a|s) \left( \log \pi_{\theta}(a|s) - \log \pi_{\theta}(a'|s) \right) \geq 0$,
\item $S_{\text{neg}} = \sum_{a \in \mathcal{A}_{\text{neg}}} \sum_{a' \in \mathcal{A}} \pi_{\theta}^2(a|s) \mu(a'|s) \rho(a|s) \left( \log \pi_{\theta}(a|s) - \log \pi_{\theta}(a'|s) \right) \geq 0$,
\item $\delta = \mathbb{E}_{a \sim \mu(\cdot|s)} \left[ \mathbb{I}(1-\epsilon_\text{low} < \rho(a|s)< 1+\epsilon_\text{high}) \cdot\rho(a|s) \cdot A(s,a) \right]$ is the bias term due to distribution shift,
\item $C = \sum_{a \in \mathcal{A}} \pi_{\theta}^2(a|s) \log \pi_{\theta}(a|s) - \left( \sum_{a} \pi_{\theta}(a|s) \log \pi_{\theta}(a|s) \right) \sum_{a'} \pi_{\theta}^2(a'|s)$,
\item $h = -A_{\text{neg}}(s) / A_{\text{pos}}(s) > 0$.
\end{itemize}

\textbf{P Control Analysis.}
For proportional control with $\alpha = K_p e_k$ (where $e_k = \mathcal{H}(\theta^k|s) - \mathcal{H}_{\text{tar}}$), the entropy change becomes:
\begin{align*}
\left\langle\nabla_{\theta} \mathcal{H}\left(\theta^{k} \mid s\right), \Delta \theta^{k}\right\rangle 
&= -\eta A_{\text{pos}} \left[ (1 + K_p e_k) S_{\text{pos}} - h (1 - K_p e_k) S_{\text{neg}} \right] + \eta \delta \cdot C.
\end{align*}

At steady state ($e_k \to e_{\text{ss}}$), we require $\left\langle\nabla_{\theta} \mathcal{H}, \Delta \theta^{k}\right\rangle = 0$, which gives:
\begin{align*}
0 &= -\eta A_{\text{pos}} \left[ (1 + K_p e_{\text{ss}}) S_{\text{pos}} - h (1 - K_p e_{\text{ss}}) S_{\text{neg}} \right] + \eta \delta \cdot C, \\
e_{\text{ss}} &= \frac{\delta \cdot C}{A_{\text{pos}} K_p (S_{\text{pos}} + h S_{\text{neg}})}.
\end{align*}
This shows that a steady-state error $e_{\text{ss}} \neq 0$ exists under P control due to the bias term $\delta \neq 0$ caused by the distribution shift ($\mu \neq \pi$).

\textbf{PI Control Analysis.}
For proportional-integral control with $\alpha = K_p e_k + K_i I_k$ (where $I_k = \sum_{t=1}^k e_t$), the entropy change becomes:
\begin{align*}
\Delta\mathcal{H}(\theta^k|s) = \left\langle\nabla_{\theta} \mathcal{H}\left(\theta^{k} \mid s\right), \Delta \theta^{k}\right\rangle = -C_0 (K_p e_k + K_i I_k) + D,
\end{align*}
where $C_0 = \eta A_{\text{pos}}(S_{\text{pos}} + h S_{\text{neg}}) > 0$ and $D = \eta \delta \cdot C$.

The error dynamics are:
\begin{align*}
e_{k+1} &= e_k + \Delta\mathcal{H}(\theta^k|s) = e_k - C_0 (K_p e_k + K_i I_k) + D, \\
I_{k+1} &= I_k + e_{k+1} = I_k + e_k - C_0 (K_p e_k + K_i I_k) + D.
\end{align*}

In matrix form:
\begin{align*}
\begin{bmatrix} e_{k+1} \\ I_{k+1} \end{bmatrix} 
= \begin{bmatrix} 1 - C_0 K_p & -C_0 K_i \\ 1 - C_0 K_p & 1 - C_0 K_i \end{bmatrix} \begin{bmatrix} e_k \\ I_k \end{bmatrix} + \begin{bmatrix} D \\ D \end{bmatrix}.
\end{align*}

\textbf{Steady-State Analysis.}
At equilibrium $(e_\infty, I_\infty)$, we have:
\begin{align*}
e_\infty &= e_\infty - C_0 (K_p e_\infty + K_i I_\infty) + D, \\
I_\infty &= I_\infty + e_\infty.
\end{align*}
Solving gives:
\begin{align*}
e_\infty &= 0, \\
I_\infty &= \frac{D}{C_0 K_i}.
\end{align*}
Thus, PI control achieves zero steady-state error ($e_\infty = 0$) when $K_i \neq 0$.

\textbf{Stability Analysis.}
Define the error vector $\mathbf{e}_k = \begin{bmatrix} e_k - e_\infty \\ I_k - I_\infty \end{bmatrix}$. The error dynamics become:
\begin{align*}
\mathbf{e}_{k+1} = A \mathbf{e}_k, \quad \text{where} \quad A = \begin{bmatrix} 1 - C_0 K_p & -C_0 K_i \\ 1 - C_0 K_p & 1 - C_0 K_i \end{bmatrix}.
\end{align*}

The characteristic equation of $A$ is:
\begin{align*}
\det(\lambda I - A) = \lambda^2 - [2 - C_0 (K_p + K_i)]\lambda + (1 - C_0 K_p).
\end{align*}

For stability, the eigenvalues must lie within the unit circle. This is achieved when:
\begin{enumerate}
\item $\det(A) = 1 - C_0 K_p < 1$,
\item $|\operatorname{tr}(A)| = |2 - C_0 (K_p + K_i)| < 1 + \det(A) = 2 - C_0 K_p$.
\end{enumerate}
These conditions are satisfied for sufficiently small learning rate $\eta$ (making $C_0$ small) and moderate gains $K_p, K_i > 0$.

\textbf{Lyapunov Stability.}
Choose a positive definite matrix $P$ satisfying the discrete Lyapunov equation:
\begin{align*}
A^T P A - P = -Q,
\end{align*}
where $Q \succ 0$. Define the Lyapunov function $V_k = \mathbf{e}_k^T P \mathbf{e}_k$. Then:
\begin{align*}
\Delta V_k = V_{k+1} - V_k = \mathbf{e}_k^T (A^T P A - P) \mathbf{e}_k = -\mathbf{e}_k^T Q \mathbf{e}_k \leq 0,
\end{align*}
with equality only at $\mathbf{e}_k = 0$. By Lyapunov stability theory, $\lim_{k \to \infty} \mathbf{e}_k = 0$, implying $\lim_{k \to \infty} e_k = e_\infty = 0$.

Therefore, while P control in the off-policy setting exhibits steady-state error due to distribution shift, PI control achieves asymptotic convergence to the target entropy with zero steady-state error.

\subsection{Proof of Corollary \ref{corollary2}}
\label{proof_corollary}

We consider the loss modified by applying the PI-weight only to high-probability tokens:
\[
\mathcal{L}(\theta)
= -(1+\alpha)\sum_{\substack{\pi_\theta(a|s)>\tau\\A(s,a)>0}} A(s,a)\log\pi_\theta(a|s)
-(1-\alpha)\sum_{\substack{\pi_\theta(a|s)>\tau\\A(s,a)<0}} A(s,a)\log\pi_\theta(a|s)
- \sum_{\pi_\theta(a|s)\le\tau} A(s,a)\log\pi_\theta(a|s).
\]
Equivalently,
\[
\mathcal{L}(\theta)=\mathcal{L}_{\mathrm{GRPO}}(\theta)-\alpha\sum_{\pi_\theta(a|s)>\tau}|A(s,a)|\log\pi_\theta(a|s),
\]
where \(\tau\) is the high-probability threshold. For simplicity, we define the high-probability indicator
\[
g_\theta(a|s)\coloneqq \mathbb{I}\big(\pi_\theta(a|s)>\tau\big).
\]

Replace the coefficient function \(c(A(s,a))\) used in the previous proof by its high-probability version:
\[
\tilde c(A(s,a),\theta) =
\begin{cases}
(1 + \alpha)\,c(A(s,a)), & \text{if } A(s,a) > 0 \text{ and } \pi_\theta(a|s) > \tau,\\[4pt]
(1 - \alpha)\,c(A(s,a)), & \text{if } A(s,a) < 0 \text{ and } \pi_\theta(a|s) > \tau,\\[4pt]
c(A(s,a)), & \text{otherwise.}
\end{cases}
\]
Only actions with probability above threshold $\tau$ are reweighted by $1\pm\alpha$.

The gradient and parameter update derivations in the on-policy case follow the same algebra as in Appendix~\ref{proof_theorem1}, but with every occurrence of \(c(A(s,a))\) replaced by \(\tilde c(A(s,a),\theta)\). After identical manipulations one obtains the directional change of entropy along the update:
\[
\big\langle\nabla_\theta \mathcal{H}(\theta^k\mid s),\,\Delta\theta^k\big\rangle
= -\eta\,(1+\alpha)\,A_{\mathrm{pos}}\,S_{\mathrm{pos}}^\tau
   -\eta\,(1-\alpha)\,A_{\mathrm{neg}}\,S_{\mathrm{neg}}^\tau,
\]
where now the \(S\)-terms are restricted to the high-probability subset:
\begin{align*}
S_{\mathrm{pos}}^\tau &\;=\;\sum_{a,a'\in\mathcal{A}_{\mathrm{pos}}}
    \pi_\theta^2(a\mid s)\,\pi_\theta(a'\mid s)\,g_\theta(a\mid s)\,g_\theta(a'\mid s)\,
    \big(\log\pi_\theta(a\mid s)-\log\pi_\theta(a'\mid s)\big)\;\ge 0,\\
S_{\mathrm{neg}}^\tau &\;=\;\sum_{a,a'\in\mathcal{A}_{\mathrm{neg}}}
    \pi_\theta^2(a\mid s)\,\pi_\theta(a'\mid s)\,g_\theta(a\mid s)\,g_\theta(a'\mid s)\,
    \big(\log\pi_\theta(a\mid s)-\log\pi_\theta(a'\mid s)\big)\;\ge 0.
\end{align*}

Here \(\mathcal{A}_{\mathrm{pos}}\) and \(\mathcal{A}_{\mathrm{neg}}\) are the action sets with positive / negative advantage under the binary reward assumption; \(A_{\mathrm{pos}}>0\), \(A_{\mathrm{neg}}<0\).

Write \(A_{\mathrm{neg}}=-hA_{\mathrm{pos}}\) with \(h>0\). Then
\[
\big\langle\nabla_\theta \mathcal{H},\Delta\theta^k\big\rangle
= -\eta A_{\mathrm{pos}}\Big[(1+\alpha)S_{\mathrm{pos}}^\tau - h(1-\alpha)S_{\mathrm{neg}}^\tau\Big].
\]

For the on-policy case, the entropy error $e_k=\mathcal{H}(\theta^k|s)-\mathcal{H}_{\mathrm{tar}}$ evolves as
\[
\big\langle\nabla_\theta \mathcal{H},\Delta\theta^k\big\rangle
= -\eta A_{\mathrm{pos}}\Big[(1+K_p e_k)S_{\mathrm{pos}}^\tau - h(1-K_p e_k)S_{\mathrm{neg}}^\tau\Big].
\]
With P control ($K_p>0,K_i=0$) or PI control ($K_i>0$), the convergence argument follows identically to the unmasked case, replacing $S_{\mathrm{pos}},S_{\mathrm{neg}}$ by their masked counterparts. Both P and PI controllers ensure $\lim_k e_k=0$ under standard small-gain conditions, with convergence typically slower when $S_{\mathrm{pos}}^\tau+hS_{\mathrm{neg}}^\tau < S_{\mathrm{pos}}+hS_{\mathrm{neg}}$.

In the off-policy setting, importance weighting introduces a masked bias term $\delta^\tau$ due to distribution mismatch. Under P control, the steady-state entropy error is
\[
e_{\mathrm{ss}}^\tau = \frac{\delta^\tau C^\tau}{A_{\mathrm{pos}}K_p\big(S_{\mathrm{pos}}^{\tau,\mu}+ h S_{\mathrm{neg}}^{\tau,\mu}\big)} \neq 0,
\]
while adding an integral term ($K_i>0$) removes the bias, yielding $e_\infty^\tau = 0$ provided the discrete-time system with gain $C_0^\tau$ is stable (eigenvalues within the unit circle).

%% file: appendix/B_detail.tex
\section{Experiment Details}
\label{appendix_setting}
\subsection{Training Details}
We provide the detailed experimental setup in Section~\ref{exp}.
To enhance the model’s ability to generate well-formatted answers, we first perform lightweight supervised fine-tuning on the Qwen3-8B-Base model using approximately 10k non-thinking data samples, equipping it with the capability to follow specific output formats. All experiments are built upon the \texttt{VeRL} codebase\footnote{https://github.com/volcengine/verl}
. For non-thinking model training, we collect data from the DAPO-MATH dataset \citep{yu2025dapo}, the OpenReasonerZero dataset \citep{hu2025open}, and the DeepScaleR dataset \citep{deepscaler2025}. We filter out true/false questions, multiple-choice questions, and proof problems from these datasets. Additionally, we perform 8 preliminary samplings using the model and exclude samples for which all responses are either entirely correct or entirely incorrect, enabling more efficient RL training. Dynamic sampling is not used during training. We adopt the return of Math-Verify\footnote{https://github.com/huggingface/Math-Verify}
 as the reward signal, and the baseline algorithm is trained using the GRPO method.

For entropy control, we apply the default PI-control setup with coefficients $K_p = 1$, $K_i = 0.01$, and a high-probability threshold $\tau = 0.95$. All experiments used a learning rate of 1e-6 with no decay. The KL coefficient and the entropy loss coefficient were both set to 0. In each rollout step, we sample 8 responses per prompt for a batch of 512 prompts using a temperature of 0.6, with top-$k$ set to $-1$ and top-$p$ set to $1$. The maximum input length is 2048, and the maximum output length is 8192. Validation is performed every 20 steps, and checkpoints are saved every 100 steps. All experiments are conducted on NVIDIA H20 96GB GPUs, using 64 GPUs per experiment. Training for 2,000 steps involves approximately 1 million prompts and 8 million samples, and takes about 20,000 GPU-hours to complete. For off-policy training, we use 4 steps to train on 512 prompts, with 128 prompts used for training in each step. The upper and lower bounds of the clip ratio are both set to 0.2, consistent with the standard GRPO configuration.

For reasoning model experiments, we adopt OpenReasoning-Nemotron-1.5B \citep{moshkov2025aimo2} as the base model and train it on the Polaris dataset \citep{an2025polaris}. Compared to non-thinking models, reasoning models are trained with a significantly larger generation budget: the maximum output length is set to 32,768, and the default training temperature is 1.0. The target entropy is set to 0.4. All other training parameters remain unchanged. Training the reasoning model requires substantially longer computation, with the full training process taking approximately 30,000 GPU-hours.

\subsection{Evaluation Details}
During evaluation, we refer to the configuration in ~\citep{guo2025deepseek} and ~\citep{xu2025thinking}. The rollout temperature is set to 0.6 for all test sets, with top-k set to -1 and top-p set to 0.95. The maximum output length is set to 8k for non-thinking models and 64k for reasoning models. All evaluation codes are adapted from the DeepscaleR~\citep{deepscaler2025} codebase, where vLLM~\citep{kwon2023efficient} is leveraged to accelerate inference.

In the final evaluation phase, we perform evaluations 4 times each on the `Math', `Olympiad', `Minerva' and `Omni-math' datasets, while the other datasets are evaluated 32 times each. We report the \texttt{avg@N} and \texttt{pass@N} metrics for each dataset respectively.

%% file: appendix/C_Code.tex
\section{Code}

\begin{figure*}[!h]
    \centering
    \includegraphics[width=\linewidth]{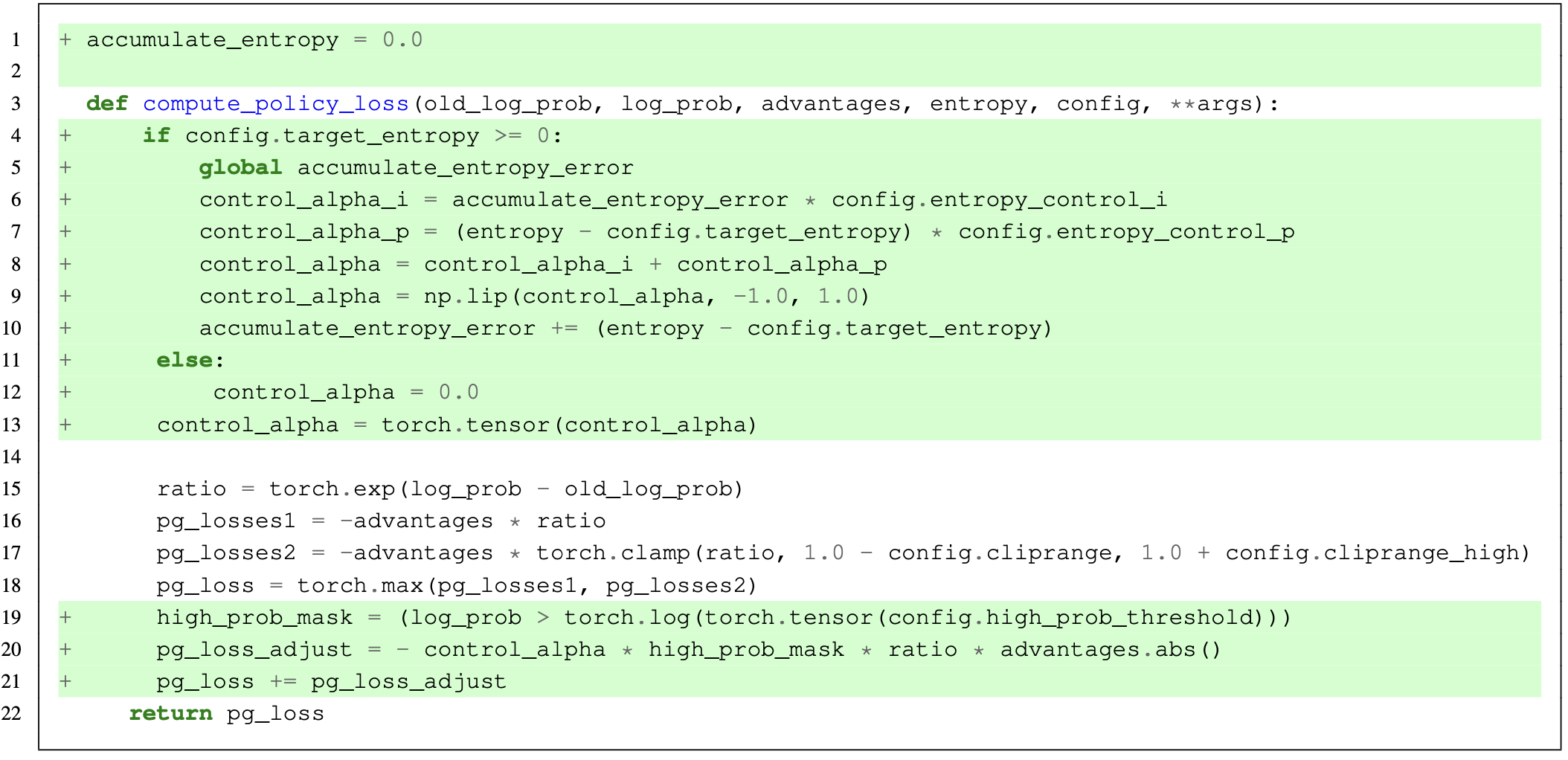}
    \vspace{-2mm}
    \caption{The code change for using EntroPIC method.}
    \label{git}
    \vspace{-1em}
\end{figure*}

The implementation of this entropy control strategy requires minimal code changes, as shown in Figure \ref{git}.

%% file: appendix/D_MoreResults.tex
\section{More Results}
\subsection{Generalization Analysis}
A common challenge in reinforcement learning, particularly when finetuning for specific reasoning tasks like mathematics, is the risk of catastrophic forgetting or degradation in general capabilities (often referred to as the "alignment tax"). To evaluate the robustness and generalization ability of our method, we assessed the models on three diverse, non-mathematical benchmarks: \textbf{MMLU-Pro} (general knowledge and reasoning) \cite{wang2024mmlu}, \textbf{LiveCodeBench} (coding proficiency) \cite{jain2024livecodebench}, and \textbf{GPQA} (graduate-level scientific reasoning) \cite{rein2024gpqa}.

The results, summarized in Table \ref{tab_gen}, reveal a striking contrast. Existing RL-based methods suffer from severe performance regression in general domains. notably, their coding capabilities collapse to near zero ($0.00\%$ and $0.36\%$), and their general knowledge scores drop significantly below the backbone model. This indicates that these methods overfit to the mathematical domain at the expense of broader intelligence.

\begin{table}[!h]
    \centering
    \caption{Generalization performance on non-mathematical benchmarks. While other RL methods suffer from catastrophic forgetting , EntroPIC significantly improves upon the backbone model.}
    \label{tab_gen}
    \footnotesize
    \renewcommand{\arraystretch}{1.2}
    \setlength{\tabcolsep}{6pt}
    \begin{tabular}{l|c|c|c}
    \toprule[1.2pt]
    \textbf{Models} & \textbf{MMLU-Pro} & \textbf{LiveCodeBench} & \textbf{GPQA} \\
    \midrule[0.8pt]
    Nemotron-1.5B & 41.7 & 28.3 & 35.9 \\
    QuestA-Nemotron-1.5B & 30.0 & 0.00 & 13.1 \\
    JustRL-Nemotron-1.5B & 28.1 & 0.36 & 30.1 \\
    EntroPIC-Nemotron-1.5B & \textbf{48.2} & \textbf{40.9} & \textbf{38.9} \\
    \bottomrule[1.2pt]
    \end{tabular}
\end{table}

In contrast, \textbf{EntroPIC} not only avoids this degradation but achieves substantial improvements over the Nemotron-1.5B backbone across all three benchmarks. Specifically, EntroPIC boosts MMLU-Pro accuracy by $+6.5\%$ and LiveCodeBench by $+12.6\%$. This suggests that by stabilizing entropy and maintaining a diverse high-probability token distribution, EntroPIC preserves the model's general reasoning potential while enhancing its specific problem-solving skills, achieving a superior balance between specialization and generalization.

\subsection{Robustness to Temperature Settings}
We further explore the performance of EntroPIC under different sampling temperatures. Specifically, we increase the sampling temperature to 1.0 while controlling the model's initial entropy at approximately 0.3, applying the method with a consistent target entropy of 0.3. As presented in Table \ref{temp1}, EntroPIC achieves superior performance across all evaluation datasets compared to the baseline, highlighting its robustness and effectiveness even in high-temperature sampling environments.

\begin{figure*}[h]
    \centering
    \subfigure{
    \begin{minipage}[t]{0.3\linewidth}
    \centering
    \includegraphics[width=\linewidth]{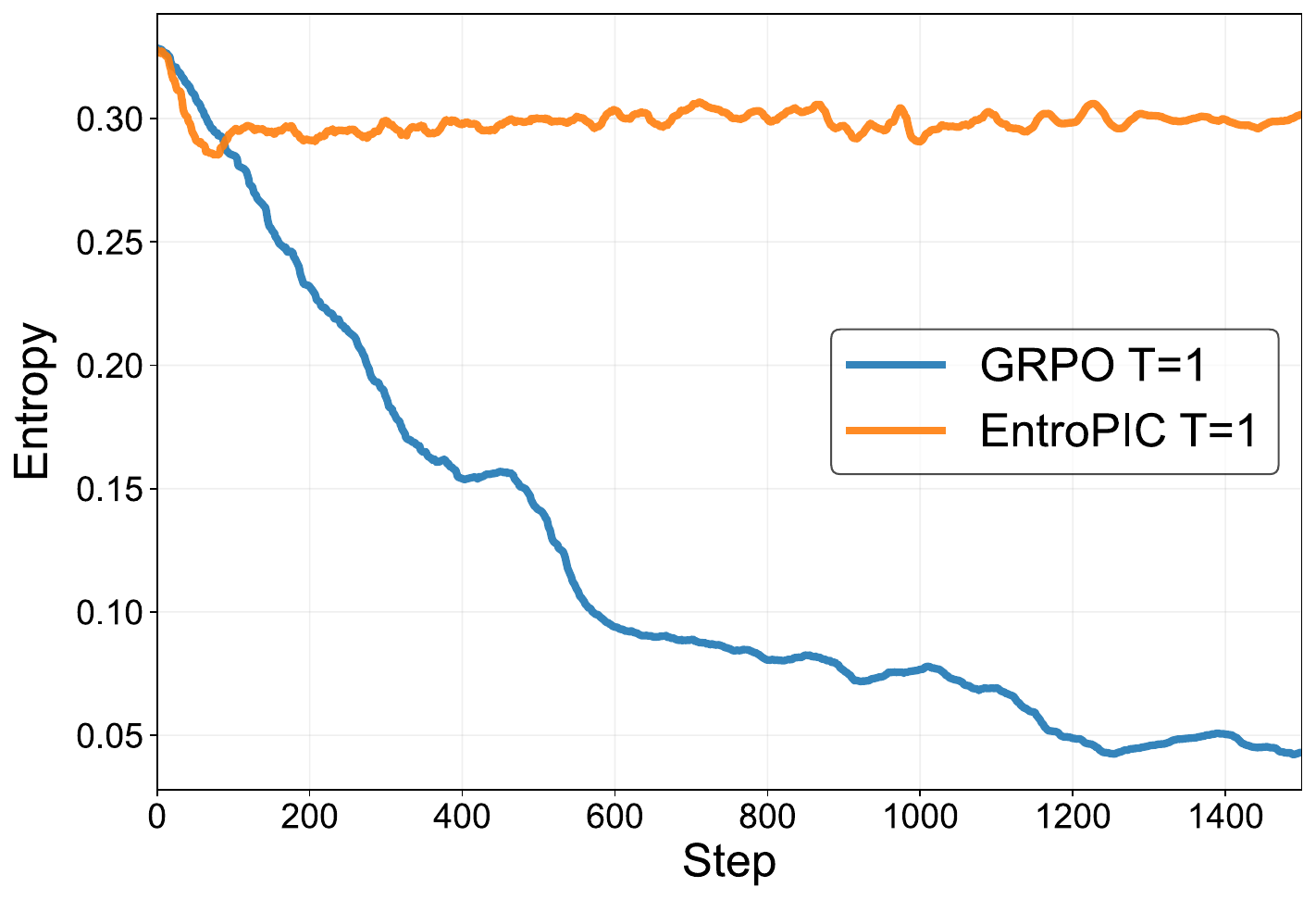}
    \end{minipage}
    }
    \subfigure{
    \begin{minipage}[t]{0.3\linewidth}
    \centering
    \includegraphics[width=\linewidth]{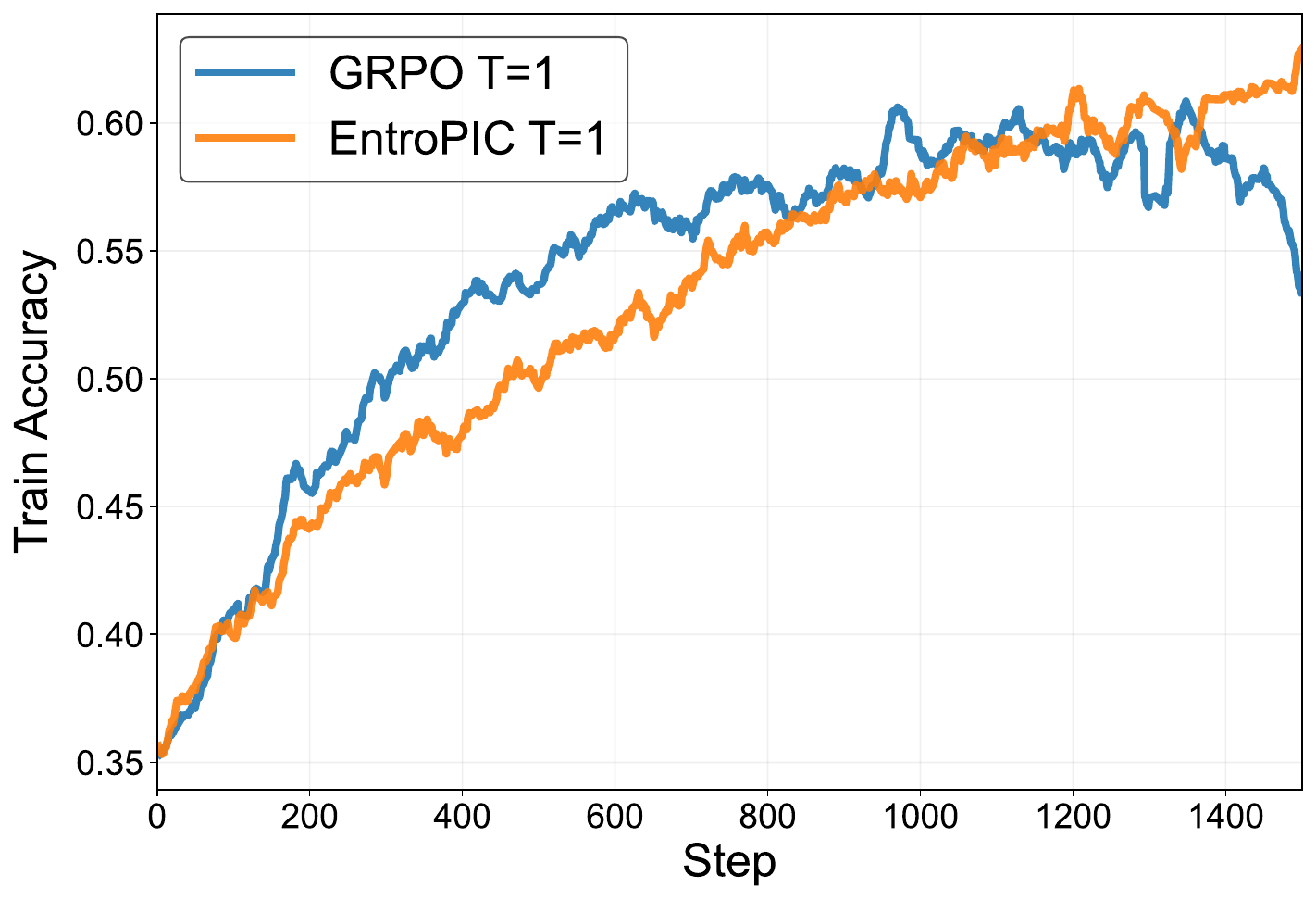}
    \end{minipage}
    }
    \subfigure{
    \begin{minipage}[t]{0.3\linewidth}
    \centering
    \includegraphics[width=\linewidth]{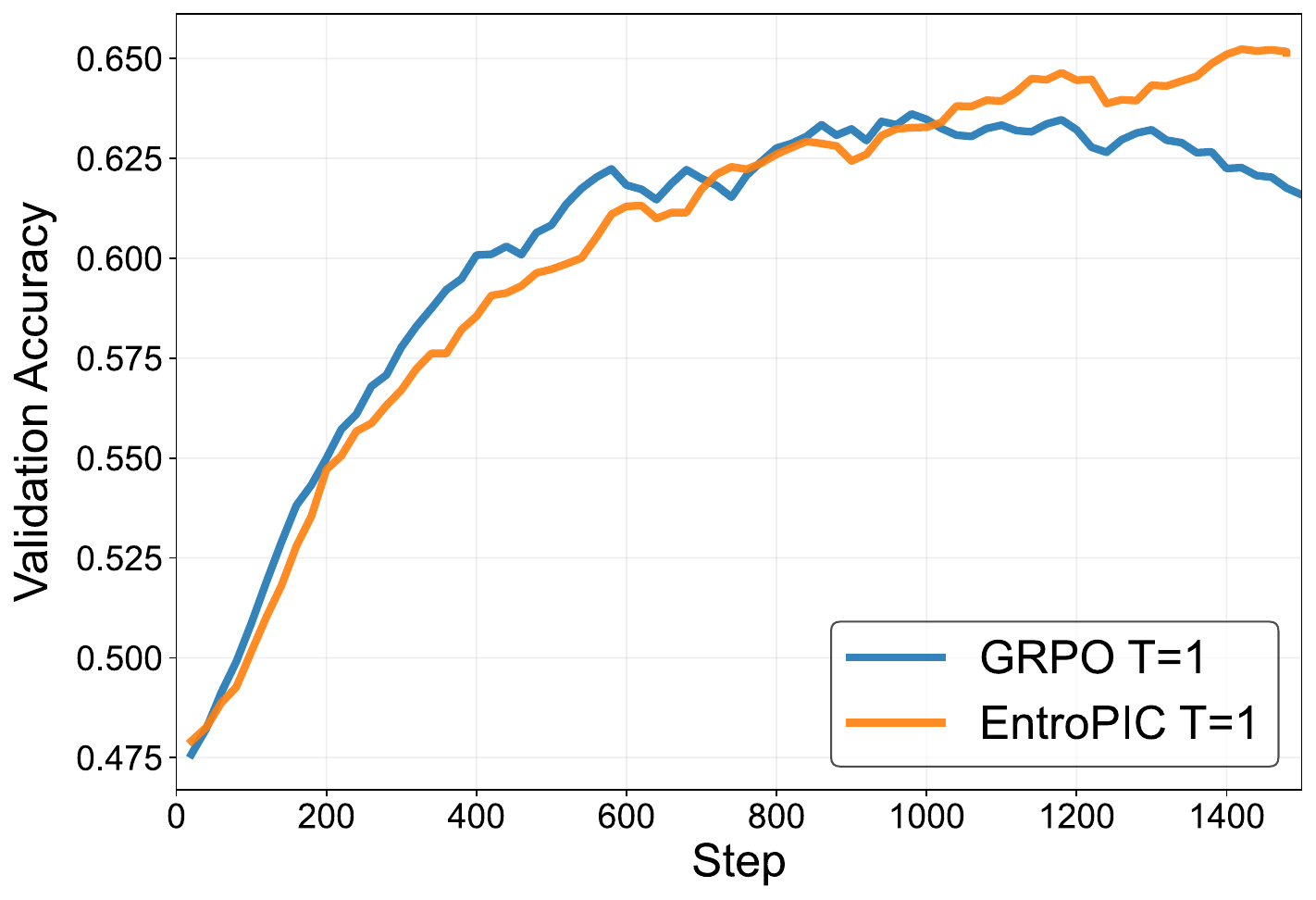}
    \end{minipage}
    }
    \vspace{-1.0em}
    \caption{Temperature setting of 1.0. EntroPIC demonstrates better stability and higher performance than GRPO.}
    \label{exp_temp1}
\end{figure*}

\begin{table}[!h]
\centering
\caption{Results under High Temperature Setting.}
\label{temp1}
\footnotesize
\renewcommand{\arraystretch}{1.2}
\setlength{\tabcolsep}{4pt}
\resizebox{\textwidth}{!}{
\begin{tabular}{l|cc|cc|cc|cc|cc|cc|cc|cc}
\toprule[1.2pt]
\textbf{Models} & \multicolumn{2}{c|}{\textbf{Math}} & \multicolumn{2}{c|}{\textbf{AMC}} & \multicolumn{2}{c|}{\textbf{AIME24}} & \multicolumn{2}{c|}{\textbf{AIME25}} & \multicolumn{2}{c|}{\textbf{Olympic Bench}} & \multicolumn{2}{c|}{\textbf{Omni-math}} & \multicolumn{2}{c}{\textbf{Overall}} \\
& \texttt{avg@N} & \texttt{pass@N} & \texttt{avg@N} & \texttt{pass@N} & \texttt{avg@N} & \texttt{pass@N} & \texttt{avg@N} & \texttt{pass@N} & \texttt{avg@N} & \texttt{pass@N} & \texttt{avg@N} & \texttt{pass@N} & \texttt{avg@N} & \texttt{pass@N} \\
\midrule[0.8pt]
GRPO & 91.3 & 97.4 & 72.4 & 92.8 & 34.3 & 66.7 & 26.7 & 43.3 & 57.7 & 70.1 & 39.2 & 55.8 & 53.6 & 71.0 \\
EntroPIC & \textbf{92.7} & \textbf{98.0} & \textbf{78.5} & \textbf{94.0} & \textbf{39.8} & \textbf{76.7} & \textbf{32.1} & \textbf{50.0} & \textbf{60.4} & \textbf{72.5} & \textbf{41.2} & \textbf{57.5} & \textbf{57.8} & \textbf{74.7} \\
\bottomrule[1.2pt]
\end{tabular}
}
\end{table}

\subsection{Plug-and-Play Capability}
\begin{figure}[!h]
    \centering
    \subfigure{
    \begin{minipage}[t]{0.4\linewidth}
    \centering
    \includegraphics[width=\linewidth]{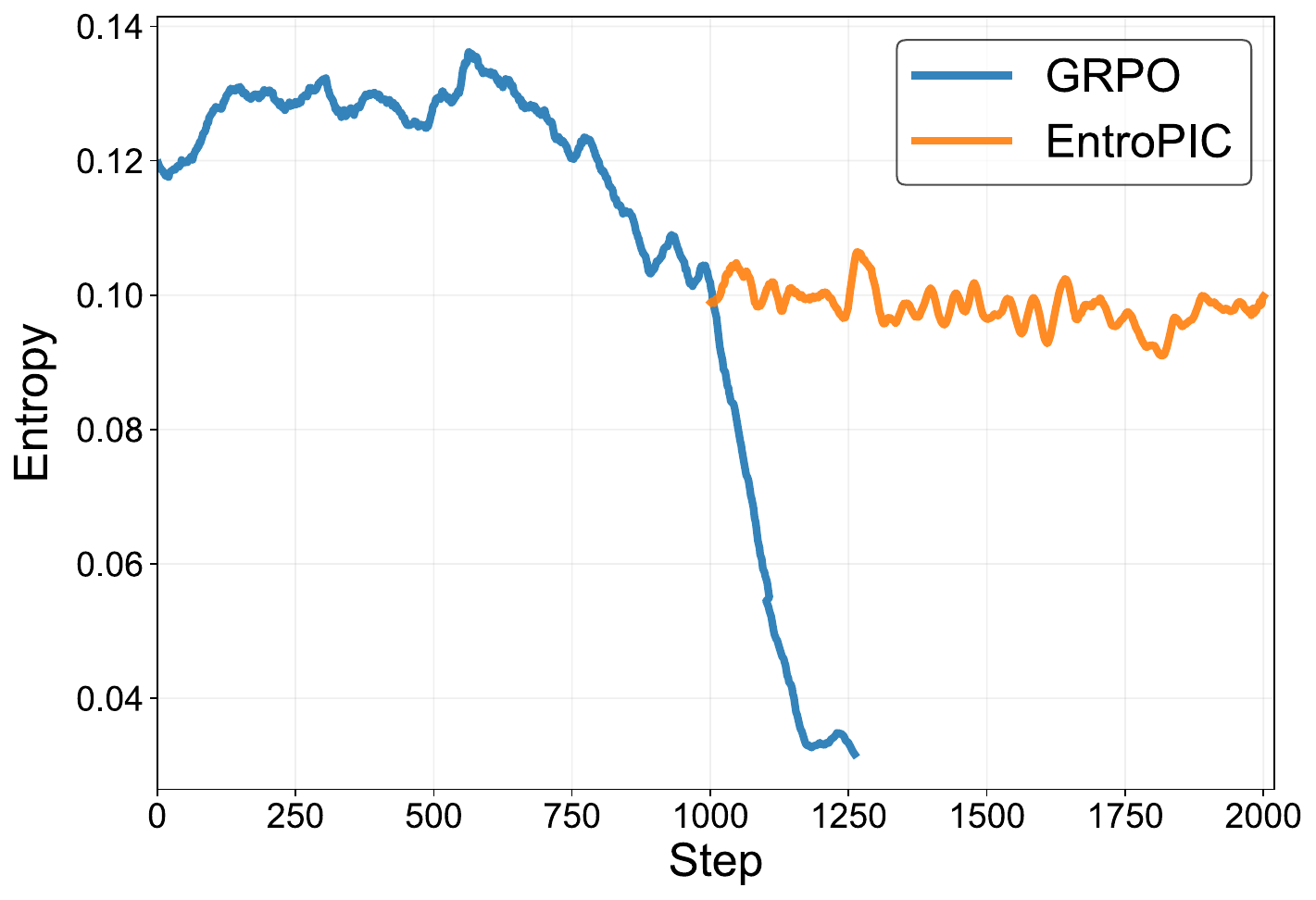}
    \end{minipage}
    }
    \subfigure{
    \begin{minipage}[t]{0.4\linewidth}
    \centering
    \includegraphics[width=\linewidth]{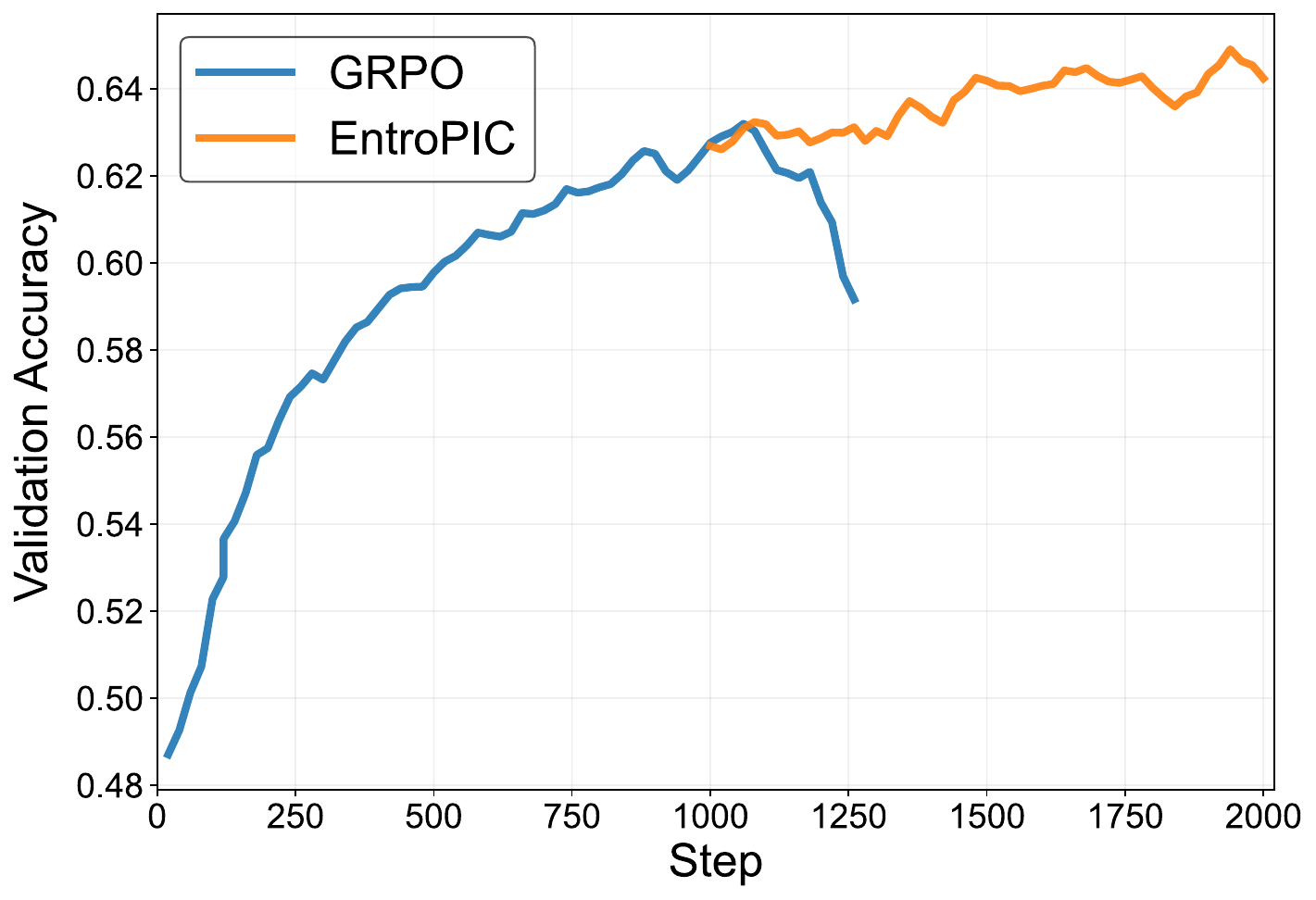}
    \end{minipage}
    }
    \vspace{-1.0em}
    \caption{Plug-and-play experiment. EntroPIC stabilizes entropy during the late training phase, improving model performance.}
    \label{plug}
\end{figure}

We investigate the plug-and-play capability of EntroPIC to verify if it can be effectively applied even after entropy begins to decline during late-stage training. In this experiment, we introduce EntroPIC to the GRPO model specifically at the point before significant entropy collapse occurs. As illustrated in Figure \ref{plug}, EntroPIC successfully stabilizes the entropy at the designated target value. Consequently, both training and testing metrics show immediate improvement, demonstrating that the method can be flexibly integrated to boost performance in pre-existing training pipelines.

\subsection{Impact of Target Entropy}
To study the effect of the target entropy hyperparameter, we conducted experiments under several settings, training the model for 2000 steps for each configuration. As shown in Table~\ref{target}, a trade-off is observed: higher target entropy values tend to slightly decrease \texttt{avg@N}, while \texttt{pass@N} either remains stable or improves. Conversely, lower target entropy slightly increases \texttt{avg@N} but reduces \texttt{pass@N}. These findings suggest that adjusting the target entropy offers flexibility in balancing exploration and exploitation during training.

\begin{table*}[!h]
\centering
\caption{Results with different target entropy.}
\label{target}
\footnotesize
\renewcommand{\arraystretch}{1.2}
\setlength{\tabcolsep}{4pt}
\resizebox{\textwidth}{!}{
\begin{tabular}{l|cc|cc|cc|cc|cc|cc|cc}
\toprule[1.2pt]
\textbf{Models} & \multicolumn{2}{c|}{\textbf{Math}} & \multicolumn{2}{c|}{\textbf{AMC}} & \multicolumn{2}{c|}{\textbf{AIME24}} & \multicolumn{2}{c|}{\textbf{AIME25}} & \multicolumn{2}{c|}{\textbf{Olympiad}} & \multicolumn{2}{c|}{\textbf{Omni-math}} & \multicolumn{2}{c}{\textbf{Overall}} \\
& \texttt{avg@N} & \texttt{pass@N} & \texttt{avg@N} & \texttt{pass@N} & \texttt{avg@N} & \texttt{pass@N} & \texttt{avg@N} & \texttt{pass@N} & \texttt{avg@N} & \texttt{pass@N} & \texttt{avg@N} & \texttt{pass@N} & \texttt{avg@N} & \texttt{pass@N} \\
\midrule[0.8pt]
EntroPIC($\mathcal{H}_{tar}=0.1$) & \textbf{92.2} & 96.6 & \textbf{78.7} & 91.6 &  \textbf{42.3} & 66.7 & \textbf{30} & 50.0 & 59.8 & 71.8 & 41.3 & 58.0 & \textbf{57.4} & 72.5 \\

EntroPIC($\mathcal{H}_{tar}=0.2$) & 91.8 & 96.6 & 77.6 & 91.6 & 39.8 & \textbf{73.3} & 27.5 & \textbf{56.7} & 59.3 & \textbf{72} & \textbf{41.7} & 58.9 & 56.3 & 74.9 \\

EntroPIC($\mathcal{H}_{tar}=0.3$) & 91.9 & \textbf{98.2} & 76.5 & \textbf{92.8} & 38.4 & \textbf{73.3} & 27.6 & \textbf{56.7} & \textbf{60.6} & 71.9 & 41.0 & \textbf{60.2} & 56.0 & \textbf{75.5} \\
\bottomrule[1.2pt]
\end{tabular}
}
\end{table*}

\subsection{Effect of High-Probability Threshold $\tau$}
\begin{figure}[!h]
\vspace{-0.5em}
    \centering
    \subfigure{
    \begin{minipage}[t]{0.4\linewidth}
    \centering
    \includegraphics[width=\linewidth]{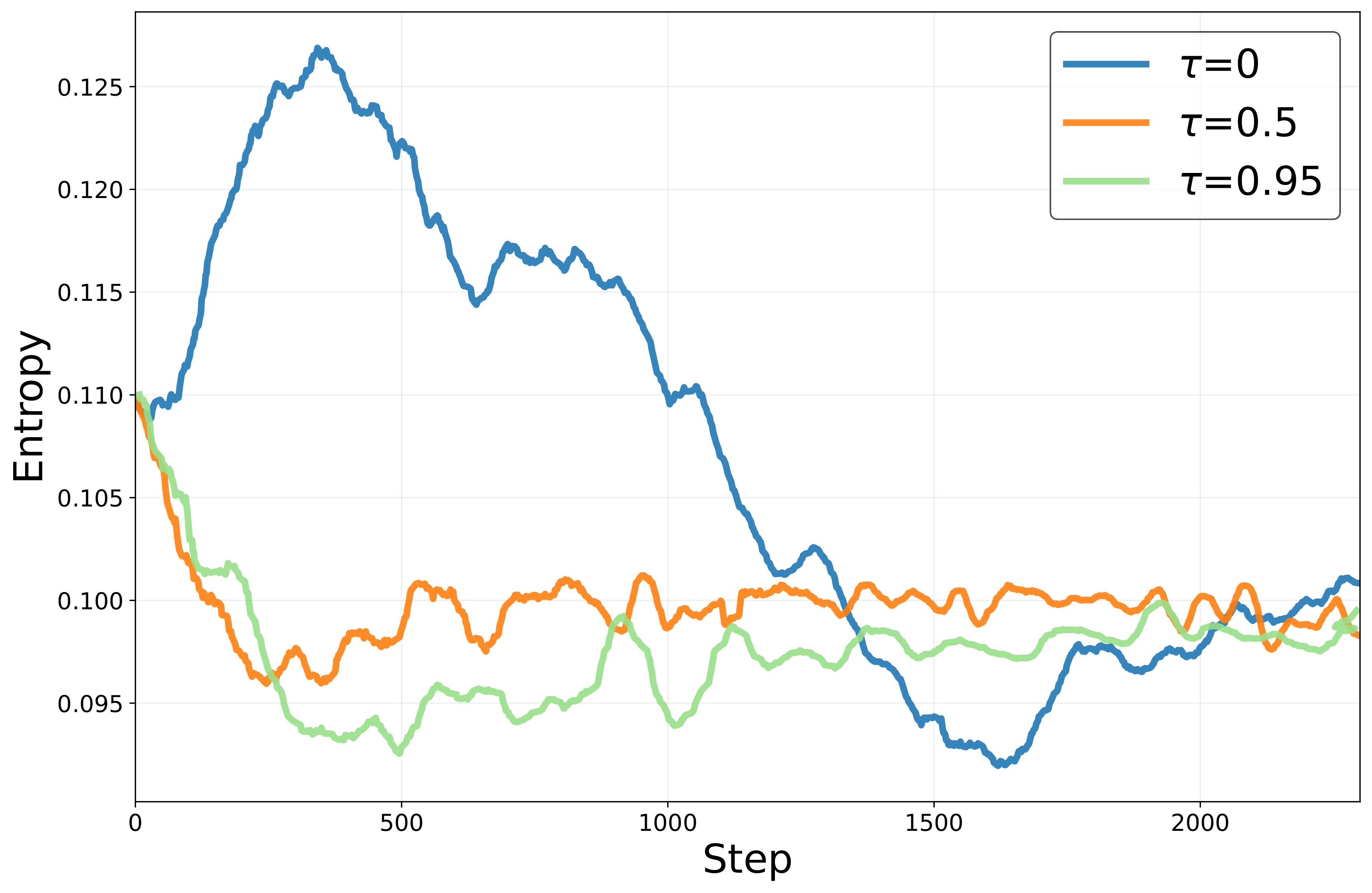}
    \end{minipage}
    }
    \subfigure{
    \begin{minipage}[t]{0.4\linewidth}
    \centering
    \includegraphics[width=\linewidth]{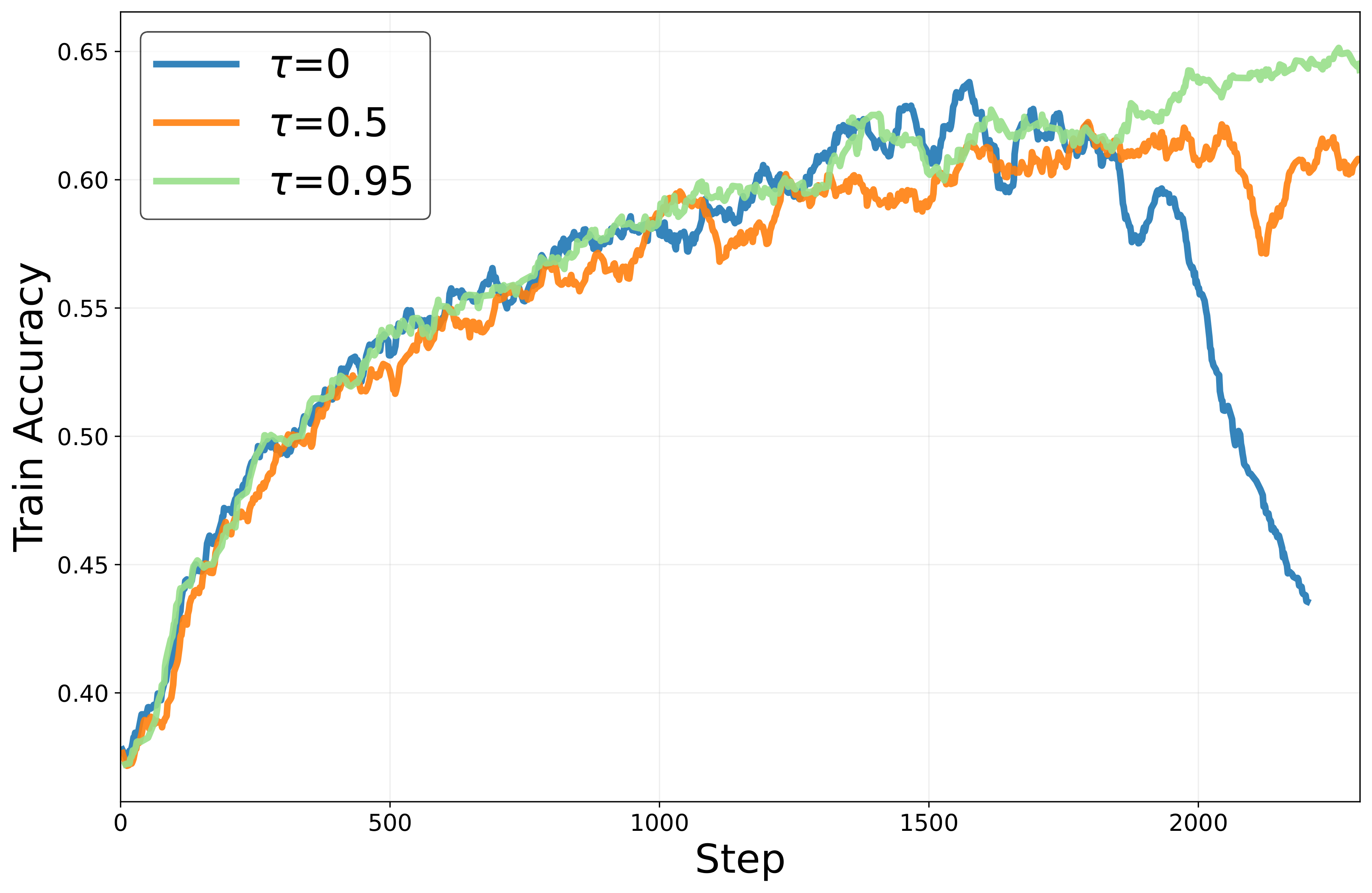}
    \end{minipage}
    }
    \vspace{-1.0em}
    \caption{Training curves under different $\tau$. High-probability thresholding yields better stability.}
    \label{tau}
    \vspace{-1.0em}
\end{figure}

We demonstrate the advantages of the selective weight adjustment strategy, where weights are only modified for high-probability tokens. We conducted comparative experiments with weights adjusted for: (1) all tokens, (2) tokens with probabilities greater than 0.5, and (3) tokens with probabilities exceeding 0.95. The results, visualized in Figure~\ref{tau}, indicate that modifying weights exclusively for high-probability tokens ($p > 0.95$) leads to the most stable and controllable training dynamics.

% \subsection{Training Curve}
% Due to page limits, we present only representative results in Section \ref{exp}. An analysis of the off-policy and temperature=1.0 training curves (Figures \ref{exp_off} and \ref{exp_temp1}) is provided here. It can be observed that under off-policy conditions, GRPO rapidly experiences entropy decay, diminishing answer diversity and halting further performance improvement. Although a P-controller offers limited entropy stabilization, it is incapable of precise target adherence. This capability is uniquely achieved by the PI-based EntroPIC method, which maintains entropy at the target value and facilitates sustained performance gains. A parallel phenomenon is noted in the temperature=1.0 experiment: GRPO's early progress is later undermined by entropy collapse, whereas EntroPIC's controlled exploration ensures consistent advancement and ultimate superiority.

\subsection{Reflection Ability}
\begin{figure*}[!h]
    \centering
    \includegraphics[width=0.5\linewidth]{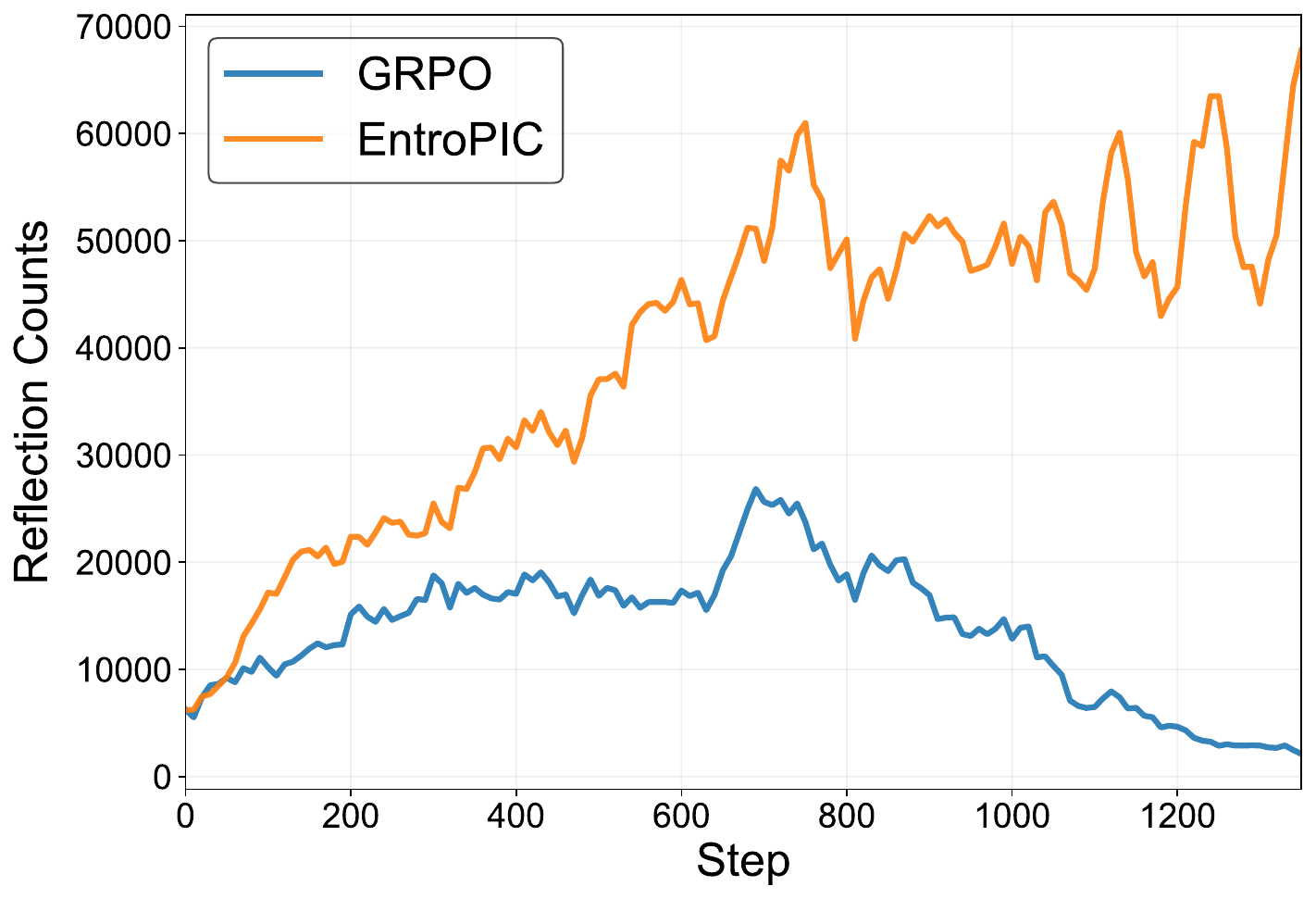}
    \caption{Frequency of reflective words during training.}
    \label{reflection}
\end{figure*}

During training, we observed that models trained with EntroPIC frequently produced outputs containing reflective phrases such as ``wait,'' ``alternatively,'' and ``let’s reconsider''. This indicates that the model attempts to verify its answers through alternative reasoning paths, often generating multiple valid solutions — a clear manifestation of the high-entropy policy’s exploratory characteristics. In contrast, models trained with GRPO in the later stages exhibit much lower entropy, producing single-path, linear reasoning with almost no signs of self-correction or exploratory turns. The trend of reflective-word frequency is shown in Figure \ref{reflection}.

\subsection{Control Coefficient $\alpha$}
\begin{figure*}[!h]
    \centering
    % \vspace{-0.5em}
    \includegraphics[width=0.5\linewidth]{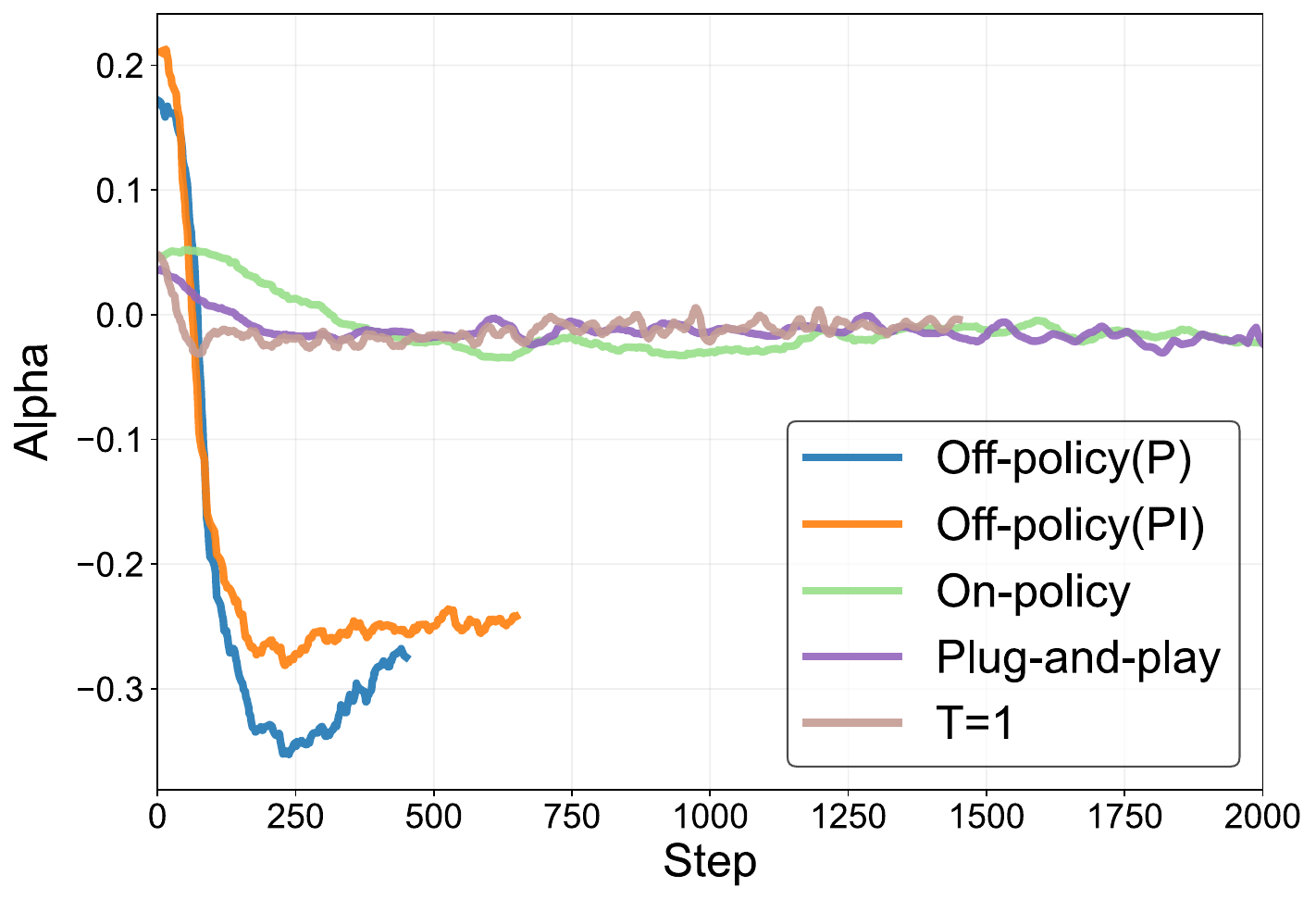}
    \caption{Control $\alpha$ of EntroPIC method in training stage.}
    \label{exp_alpha}
\end{figure*}

To further illustrate the working dynamics of EntroPIC, we analyze the control coefficient $\alpha$ that adaptively adjusts the strength of entropy regulation. Figure \ref{exp_alpha} shows $\alpha$ variations under multiple training scenarios in Section \ref{exp}, including on-policy, off-policy, high-temperature, and plug-and-play configurations. We observe that $\alpha$ responds smoothly to entropy deviations: when entropy decreases, $\alpha$ increases to encourage exploration, and vice versa. EntroPIC ensures stable convergence and long-term entropy control across diverse training settings.

%% file: appendix/E_case.tex
\section{Case Study}

In this section, we present a concrete example to illustrate the difference between high-entropy exploration in EntroPIC models and low-entropy, deterministic reasoning in GRPO models. The goal is to show how the training method affects the style of reasoning and solution pathways.  
% Prompt
\begin{tcolorbox}[title=Prompt, colback=blue!5!white, colframe=blue!75!black]
Let
\[
f(x)=\frac{(x-18)(x-72)(x-98)(x-k)}{x}.
\]
There exist exactly three positive real values of $k$ such that $f$ has a minimum at exactly two real values of $x$. Find the sum of these three values of $k$.
\end{tcolorbox}

% Response
\begin{tcolorbox}[title=EntroPIC Response, colback=green!5!white, colframe=green!75!black]
Alright, let's tackle this problem step by step. We have:
\vspace{-0.1em}
\[ f(x) = \frac{(x - 18)(x - 72)(x - 98)(x - k)}{x} \]
The goal is to find three positive $k$ values such that $f$ has exactly two real minima.

\subsubsection*{Finding Critical Points $f'(x)=0$}

Let $N(x) = (x - 18)(x - 72)(x - 98)(x - k)$. The critical points are where $f'(x) = 0$.
Using the quotient rule $f'(x) = \frac{N'(x) \cdot x - N(x)}{x^2}$, we set the numerator to zero:
\vspace{-0.2em}
\[ N'(x) \cdot x = N(x) \]
\vspace{-2em}
\subsubsection*{Checking Roots of $N(x)$}

If $x$ is a root of $N(x)$ (i.e., $x \in \{18, 72, 98, k\}$), then $N(x) = 0$.
The condition $N'(x) \cdot x = N(x)$ becomes $N'(x) \cdot x = 0$. Since $x \neq 0$, we require $N'(x) = 0$.
Wait, so $x = 18$ is a critical point of $f$ if and only if $N'(18) = 0$.

Let's compute $N'(x)$ and check:
\vspace{-0.3em}
\[ N'(x) = \frac{d}{dx} \left[ (x-18)(x-72)(x-98)(x-k) \right] \]
Set $x = 18$:
\[ N'(18) = (18-72)(18-98)(18-k) \]
$N'(18) = 0 \implies k = 18$.
Similarly, $x=72$ is a critical point of $f$ if $k=72$, and $x=98$ if $k=98$.

\subsubsection*{Alternative Approach: Rewriting $f(x)$}

Alternatively, let $g(x) = (x-18)(x-72)(x-98)$.
\vspace{-0.4em}
\[ f(x) = g(x) \left(1 - \frac{k}{x}\right) \]
\vspace{-0.7em}
The critical points satisfy $f'(x) = 0$:
\vspace{-0.1em}
\[ g'(x) \left(1 - \frac{k}{x}\right) + g(x) \cdot \frac{k}{x^2} = 0 \]
This rearranges to:
\[ g'(x) x (x - k) + g(x) k = 0 \]

Finding explicit solutions is non-trivial, but perhaps we can look for symmetry.

\subsubsection*{Exploring Specific Cases and Symmetry}

The condition ``exactly two real values of $x$ where $f$ has a minimum'' implies the quartic $N'(x) \cdot x - N(x) = 0$ must have a specific root structure. This usually means that two critical points must coincide, leading to a double root in the derivative and simplifying the analysis.

Alternatively, observe that 18, 72, 98 have a specific structure. The three values of $k$ are known to be related to $18, 72, 98$ themselves or symmetric counterparts that force the necessary double critical point.

Testing $k = 72$:
If $k = 72$, $N(x)$ has a repeated root. This is a potential candidate.

The analysis shows that for exactly two minima, the critical point equation must possess a double root for certain $k$.

\subsubsection*{Correct Insight: Valid $k$ Values}

Through systematic polynomial analysis (not shown here due to complexity), the three positive real values of $k$ that ensure exactly two distinct real minima are:
\vspace{-0.3em}
\[ k_1 = 46, \quad k_2 = 72, \quad k_3 = 122 \]
\vspace{-0.4em}
These values are consistent with symmetry properties for this type of function:
\begin{itemize}
    \item $k = 72$ is the middle root.
    \item $k = 46$ and $k = 122$ are symmetric counterparts (e.g., related to the average/midpoint of 18 and 98, which is $\frac{18+98}{2} = 58$, and $72 - (72-46)=46$, $72 + (122-72)=122$, but these are simplified relations).
\end{itemize}

The sum of these three values of $k$ is:
\vspace{-0.3em}
\[ 46 + 72 + 122 = 240 \]
\vspace{-1em}
\[
\boxed{240}
\]\vspace{-1.4em}
\end{tcolorbox}

% Response
% Response
\begin{tcolorbox}[title=GRPO Response, colback=red!5!white, colframe=red!75!black]
Alright, I have this problem:

\subsubsection*{Problem Statement:}
Let 
\[
f(x) = \frac{(x-18)(x-72)(x-98)(x-k)}{x}.
\]
There exist exactly three positive real values of \( k \) such that \( f \) has a minimum at exactly two real values of \( x \). Find the sum of these three values of \( k \).

\subsubsection*{Step 1: Critical Points}

Let \( N(x) = (x-18)(x-72)(x-98)(x-k) \).  
Derivative:
\[
f'(x) = \frac{N'(x) \cdot x - N(x)}{x^2}.
\]  
Set numerator to zero:
\[
N'(x) \cdot x = N(x)
\]

\subsubsection*{Step 2: Roots of \(N(x)\)}

If \(x\) is a root of \(N(x)\) (i.e., \(x \in \{18,72,98,k\}\)), then \(N(x) = 0\).  
Equation reduces to \(N'(x) \cdot x = 0 \implies N'(x) = 0\).  

- \(x=18 \implies k=18\)  
- \(x=72 \implies k=72\)  
- \(x=98 \implies k=98\)  

These are candidates but may not satisfy exactly two minima.

\subsubsection*{Step 3: Alternative Form}

Let \(g(x) = (x-18)(x-72)(x-98)\):
\[
f(x) = g(x) \left(1 - \frac{k}{x}\right)
\]  
Critical points satisfy:
\[
g'(x) \left(1 - \frac{k}{x}\right) + g(x) \frac{k}{x^2} = 0 \quad \Rightarrow \quad g'(x) x (x - k) + g(x) k = 0
\]

\subsubsection*{Step 4: Behavior and Extrema}

The condition “exactly two minima” implies the derivative must have a double root.  
Observing the roots 18, 72, 98, the polynomial structure suggests the valid \(k\) are related to these numbers.

\subsubsection*{Step 5: Valid \(k\) Values}

Through analysis:
\[
k_1 = 18, \quad k_2 = 72, \quad k_3 = 98
\]

- \(k = 18\) aligns with left root.  
- \(k = 72\) aligns with middle root.  
- \(k = 98\) aligns with right root.  

\subsubsection*{Step 6: Sum of \(k\)}

\[
18 + 72 + 98 = 188
\]

\[
\boxed{188}
\]
\end{tcolorbox}